# Online Two-stage Thermal History Prediction Method for Metal Additive Manufacturing of Thin Walls

**Yifan Tang** [1]
Email: yta88@sfu.ca
**M. Rahmani Dehaghani** [1]
Email: mra91@sfu.ca
**Pouyan Sajadi** [1]
Email: sps11@sfu.ca
**Shahriar Bakrani Balani** [2]
Email: Shahriar.bakranibalani@tuni.fi
**Akshay Dhalpe** [2]
Email: akshay.dhalpe@tuni.fi
**Suraj Panicker** [2]
Email: suraj.panicker@tuni.fi
**Di Wu** [2]
Email: di.wu@tuni.fi
**Eric Coatanea** [2]
Email: eric.coatanea@tuni.fi
**G. Gary Wang** [1*]
Email: gary_wang@sfu.ca

[1] Product Design and Optimization Laboratory, Simon Fraser University, Surrey, BC, Canada
[2] Faculty of Engineering and Natural Sciences, Automation Technology and Mechanical Engineering, Tampere University, Tampere, 33720, Finland

## ABSTRACT

Various data-driven modeling methods have been developed to predict the thermal history in metal additive manufacturing (AM) using simulation data. Their generalization capability is often unacceptable for online applications (e.g., *in-situ* control) of metal AM since the training data are only from simulation, which is usually based on simplified assumptions of the complex physics of the actual metal printing. This paper aims to propose an online two-stage thermal history prediction method, which could be integrated into a metal AM process for performance control. Based on the similarity of temperature curves (curve segments of a temperature profile of one point) between any two successive layers, the first stage of the proposed method designs a layer-to-layer prediction model (artificial neural network) to estimate the temperature curves of the yet-to-print layer from measured temperatures of certain points on the previously printed layer. With measured/predicted temperature profiles of several points on the same layer, the second stage proposes a reduced order model (ROM) (intra-layer prediction model) to decompose and construct the temperature profiles of all points on the same layer, which could be used to build the temperature field of the entire layer. The training of ROM is performed with an extreme learning machine (ELM) for computational efficiency. Fifteen wire arc AM experiments and nine simulations are designed for thin walls with a fixed length and unidirectional printing of each layer. The test results indicate that the proposed prediction method could construct the thermal history of a yet-to-print layer within 0.1 seconds on a low-cost desktop computer (Intel Core i7-3770 CPU @ 3.40GHz processor, 24.0 GB RAM). Meanwhile, the method has acceptable generalization capability in most cases from lower layers to higher layers in the same simulation, as well as from one simulation to a new simulation on different AM process parameters. More importantly, after fine-tuning the proposed method with limited experimental data, the relative errors of all predicted temperature profiles on a new experiment are smaller than 0.09, which demonstrates the applicability and generalization of the proposed two-stage thermal history prediction method in online applications for metal AM.

## 1 Introduction

In metal additive manufacturing (AM), the metallic material (e.g., powder or wire) is heated by the energy source (e.g., laser, electron beam, and electric arc) and melted to liquid, after which it cools down and solidifies. Then the deposited material is re-heated by the heat input on the higher layers and then cools down, which is repeated until the part is printed completely. The thermal behavior encountered during printing affects the final properties of the part significantly. For instance, a larger thermal gradient would bring unacceptable residual stresses and thermal distortion, resulting in part failure [1], while a smaller thermal gradient and cooling rate could reduce the hardness of a part [2].

* Corresponding Author.

To maintain stable part qualities in metal AM, various data-driven methods have been developed to predict and control the thermal history during printing. Based on data collected from finite element analysis (FEA) simulations, the data-driven methods select a machine learning technique to learn the relationship between thermal history and various parameters of interest. Compared with physics-based methods, it is more efficient computationally and requires less knowledge about the physics behind the metal AM process. According to the parameters selected as input variables, the applications of data-driven methods could be classified as those without and those with thermal information input.

### 1.1 Applications without thermal information input

These applications only consider features extracted from process settings and geometries as input variables, while the thermal effects among elements in the metal AM part is not used in the modeling methods.

With 250,000 data from FEA simulations for the direct energy deposition (DED) process of stainless steel 316L, Mozaffar *et al*. [3] trained a recurrent neural network for thermal history prediction. In the model, the input variables were constructed as a time series structure, containing the distance-based toolpath feature, the deposition time, the distance to the part boundary, layer height, laser intensity, and laser state. After training, the mean square error of the prediction was smaller than $1e^{-4}$ for FEA simulations with different laser powers, scan speeds, toolpath strategies, and geometries. However, the part boundary would vary significantly in complex geometries, which decreases the generalization performance of some distance-based input variables.

For the DED process with a composite coating (316L and tungsten carbides), Fetni *et al*. [4] selected an artificial neural network to predict the thermal history from seven input features (e.g., point location, time, laser position, distance between the laser head and point, etc.). By extending the above input features with the input energy and the layer number, Pham *et al*. [5] utilized a feed-forward neural network for the DED process with M4 high-speed steel material powder. Both works collected training data from two-dimensional FEA simulations, and their prediction accuracies of thermal histories on selected points were over 99%. However, their training relies on huge data sets, i.e., 4.1 million [4], and 19.9 million [5]. When only limited data is accessible, both models would suffer from unacceptable prediction performances caused by the data insufficiency. Moreover, as their models exclude process parameters in the input variables, both trained models cannot be applied directly to another new AM part with a different printing setting, which indicates a low generalization performance.

Based on the thermal physics, Ness *et al*. [6] designed eight generic features (e.g., sample time, deposition time, Euclidean distance, deposition status of adjacent nodes, power influence, etc.) for wire arc additive manufacturing (WAAM) processes with various geometries, deposition patterns and power intensities. The relationship between the designed features and the thermal history was learned by the extremely randomized tree model, which was trained by over 427,718 pieces of data collected from ABAQUS. Compared with temperature profiles from FEA simulations, the trained model provided predictions with a mean absolute percentage error below 10% in most cases for new WAAM parts.

To identify the effects of laser scanning patterns, Ren *et al*. [7] proposed a deep recurrent neural network model for laser aided AM process. A two-dimensional laser deposition status matrix was designed as the input to reflect the scanning path, and the output temperature field was also represented as a matrix. After training with 47,152 pieces of data collected from FEA simulations, the model provided a prediction accuracy of over 95% for various AM parts with any arbitrary geometry, but only parts with one single layer were discussed.

Different from the above methods relying on simulations only, Liao *et al*. [8] applied the physics-informed neural network (PINN) to design a hybrid physics-based data-driven thermal modeling approach for the DED process. The partial differential equation (PDE) of heat transfer was solved by the PINN model, whose training loss contained the PDE-based loss and the data-based loss, i.e., the mean square error between the prediction and the simulation/experimental data. Based on partial experimental temperature data captured by a coaxial IR camera, the trained model could predict the full temperature field of the thin wall with a root mean square error of 47.28 K, around 2.4% of the maximum experimental temperature. However, the time-consuming training process (e.g., 100 thousand epochs in the experimental study) restricts its applicability in online updates with data measured during printing.

### 1.2 Applications with thermal information input

Different from the above applications, those applications with thermal information input use the temperature of adjacent elements or the temperature of previous time steps as complementary input variables, to represent the thermal transfer effect.

Paul *et al*. [9] proposed an iterative prediction flow based on the extremely randomized tree model, which was trained on data from time-dependent heat equations. Given one element, its temperature at one time was predicted from its temperatures during the last five time steps, temperatures of surrounding elements at the previous time step, the relative location of the element to the laser input, and two time-related parameters. This iterative flow could predict the temperature profile of one element for the future 1000 time-steps with an R-square value of 0.969 and a relative mean absolute error below 1%. This method is promising to be integrated into a model-based real-time AM control system, as the prediction accuracy for multiple future time steps is high. However, the computational time could be a limitation, as 68.69 seconds were required to predict 200 future time steps.

To consider heat transfer effects between layers in gas metal arc welding based AM, Zhou *et al*. [10] designed one temperature state matrix as the supplementary input for the method [7] whose input is only the deposition order state matrix. Each element in the temperature state is the average temperature value of adjacent nodes. After training with 98,585 FEA simulation data, the model could predict the thermal history of multi-layer parts with an accuracy above 94% on the testing data. However, the small size of the predefined input and output matrices would restrict the generalization performance of the pretrained model on AM parts with larger geometry sizes.

Given any arbitrary laser trajectory in the laser-based AM, Stathatos and Vosniakos [11] proposed a distance-based decomposition method to obtain the descriptor for each trajectory point. Then the temperature of one point was predicted by artificial neural networks from the point descriptor and its temperatures at previous time steps. Based on data collected from 450 FEA simulations, the trained model could predict the thermal history of various tracks (random walk, hatch pattern, etc.) with max relative errors smaller than 5% and execution time smaller than half of the real process time. However, only a single layer was considered, and the generalization performance to multi-layer parts was not discussed.

To improve the generalizability of data-driven models in different geometries, Mozaffar *et al*. [12] selected the graph neural network (GNN) to capture the spatiotemporal dependencies of elements in the DED part. The thermal responses at one time step could be predicted by GNN from the thermal responses and the process parameters in the previous time step. The thermal responses for several future time steps could be estimated by the designed recurrent GNN, where the output from one time-step was the input for the next time step. With the dataset generated from 50 simulated DED parts with different geometries, the trained model can predict the thermal history for a new geometry with a root mean square error smaller than 0.02.

### 1.3 Summary remarks

All the above data-driven modeling methods perform well on simulations, but several common limitations exist as follows:

(a) *Few considered experimental data*. Only data from FEA simulations or time-dependent heat transfer equations are chosen for training and testing, while their performances on actual metal AM printing (i.e., experiments) are seldom discussed and validated, except for [8].

(b) *Low transferability from simulation to experiments*. Different from FEA simulations and partial differential equations, the thermal behavior in actual metal AM processes is more complex as the uncertainties in process parameters and geometry deviations would affect the thermal behavior. Therefore, large prediction errors are anticipated if these models are applied to physical experiments directly.

(c) *Limited applicability in online applications*. Considering the possible unacceptable prediction performance in actual metal AM, the above models could not be applied for online applications, such as *in-situ* control based on predicted temperature.

Different from predicting the thermal history offline, the purpose of this paper is to provide an online thermal history prediction method for metal AM processes, which could be integrated into the machine controller for process control. The key assumption is that some pyrometers or thermal cameras are available to detect the thermal history of printed parts during printing. Then the thermal history of the yet-to-print layer could be estimated online from the detected data of lower layers by the proposed method. This provides a possibility to control the dwell time of higher layers and modify the printing settings in advance if the thermal history of higher layers is not the desired one. Besides, if the relationship between thermal history and part/structure properties is modeled by some methods, the proposed method could be integrated to control the quality online.

The remainder of this paper is structured as follows. Section 2 discusses two unique characteristics of thermal behavior in metal AM, i.e., curve similarity between successive layers, and profile similarity within one layer. Section 3.1 describes the detailed settings of WAAM experiments and simulations in COMSOL Multiphysics software, as well as profile preprocessing methods. Section 3.2 presents the structure of the proposed online two-stage prediction method, including the layer-to-layer prediction model for curve similarity, and the intra-layer prediction model for profile similarity. The validation, and test performances of the proposed method are shown in Section 4. Finally, Section 5 gives a summary and some discussions for future work.

## 2 Characteristics of Thermal Behavior

The additive and repetitive nature of a metal AM process brings about similar thermal behavior to all points in one part [13]. Moreover, the thermal behavior similarity is determined by the same fundamental physics (i.e., the continuous heat diffusion equation) followed by all points. The major difference among points is the different initial conditions caused by the thermal accumulation during printing. From the perspective of layers, the similarity could be categorized as the thermal behavior similarity within one layer and between layers, which are discussed in Section 2.1 and Section 2.2 respectively. In both sections, all data are collected from a metal thin wall simulated with unidirectional printing using COMSOL Multiphysics, where a fixed travel speed is used and the dwell time is controlled to make the interpass temperature around 200 °C for each layer [2].

### 2.1 Thermal profiles within one layer

Figure 1 depicts a thin wall (left most) with the thermal profiles of three points on the $i$-th layer under both global and local time schemes. The "*global time scheme*" is defined based on the time after the metal AM process starts working, while the "*local time scheme*" for one point is defined based on the time after the element at this point is activated.

When starting to print the $i$-th layer, the material is deposited to the boundary point $p_{0,i}$ and the element is activated in the simulation model. Then the temperature of $p_{0,i}$ is recorded during further printing, as shown in the temperature profile of $p_{0,i}$ in Figure 1 (a). If the time when AM process starts working is defined as 0, the global time $t_{0,i}$ when activating $p_{0,i}$ is equal to the sum of total printing time and total dwell time of all former $i$-1 layers. When the nozzle moves a certain distance $d_j$, the corresponding element at the point $p_{j,i}$ is activated at time $t_{j,i}$, after which its temperature is measured as shown in the second profile in Figure 1 (a). Considering a fixed travel speed ($TS$) is applied, the "*relative delay*" of the point $p_{j,i}$ on the $i$-th layer is defined as $t_{j,i} - t_{0,i} = d_j/TS$, which could be used to differentiate various points on the same layer. Based on this logic, all points on the $i$-th layer would be activated chronologically until the last boundary point $p_{end,i}$ on the layer.

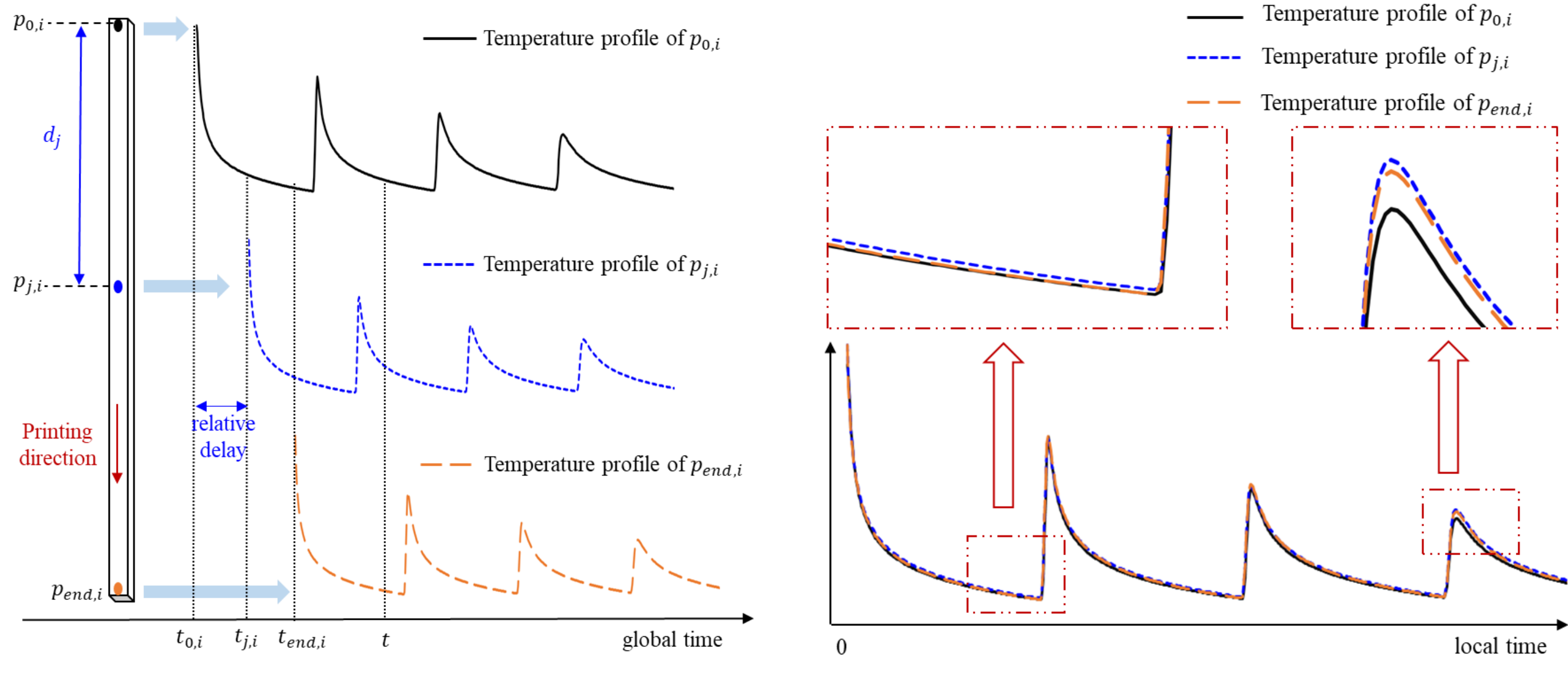

(a) Profiles under the global time scheme

(b) Profiles under the local time scheme

Figure 1 Similarity of thermal profiles within the $i$-th layer

After the $i$-th layer is printed completely, the dwell time $t^i_{dwell}$ is applied to cool the layer down to around the required interpass temperature. Meanwhile, the temperatures of all points on the layer decrease gradually. When printing the ($i$+1)-th layer, the temperature at the boundary point $p_{0,i}$ on the $i$-th layer would increase sharply as shown as the second peak on its profile in Figure 1 (a). The reason is that its element would be re-melted or re-heated by the heat input on the next layer. But the temperatures of points on the later part of the $i$-th layer keep reducing, as the heat input is far from these points. With the nozzle moving, the temperature at different points would increase to the peak gradually until finishing the ($i$+1)-th layer. Such re-heating and re-cooling processes would repeat when printing higher layers until the entire AM part is completed.

At any global time $t$, the temperature field of the $i$-th layer consists of temperatures of all points on the layer. If the temperature profiles of all points are shifted to their local time scheme, two features could be observed from Figure 1 (b). First, the temperature trends of different profiles are similar at the same local time, while their temperature values differ**.** More specifically, the points on the same layer would cool down to different temperatures or re-heat to various peaks at the same local time. Second, the time duration between two successive peaks or two lowest temperatures is similar for different points. For example, the time differences from the first peak to the second peak on three temperature profiles are close for points $p_{0,i}$, $p_{j,i}$, and $p_{end,i}$, whose durations from the first peak to the first lowest value are also similar. Therefore, temperature profiles of points on the same layer have some similarities in terms of temperature trends, as well as re-heating and re-cooling durations. The similarity observed in profiles of points on the same layer is defined as "*profile similarity*" in further discussions.

## 2.2 Thermal profiles between layers

In this paper, the point $p_{j,i}$ on the $i$-th layer and the point $p_{j,i+1}$ on the ($i$+1)-th layer are defined as one "*point pair*" denoted as $(p_{j,i}, p_{j,i+1})$, whose temperature profiles are shown in Figure 2 (b). Both points have the same distance $d_j$ to the boundary point of their corresponding layer, as shown in Figure 2 (a).

When the material is deposited to $p_{j,i}$ at $t_{j,i}$, its temperature would jump to the maximum value after which it cools down until printing the next layer. Before deposing material to $p_{j,i+1}$, the temperature at $p_{j,i}$ reduces and then increases as the heat input moves closer to the point gradually. At the time $t_{j,i+1}$, the element at the point $p_{j,i+1}$ is activated and its initial temperature is the max value. Meanwhile, the temperature at $p_{j,i}$ would increase to the peak, as the material at $p_{j,i}$ could be re-melted. From $t_{j,i+1}$ to the time $t_{j,i+2}$ when the material is deposited to the point on the ($i$+2)-th layer above $p_{j,i+1}$, the elements of $p_{j,i}$ and $p_{j,i+1}$ cool down and then are re-heated. More importantly, the temperature curve of the point $p_{j,i+1}$ during $[t_{j,i+1}, t_{j,i+2}]$ behaves similarly to the curve of the point $p_{j,i}$ during $[t_{j,i}, t_{j,i+1}]$, as shown in the first curves in Figure 2 (b). This similarity is attributed to the same phase "*activating-cooling-heating*" encountered by both points. Apart from the different initial conditions caused by heat accumulation, the major difference on both curves is the time duration. As a fixed travel speed is applied and different dwell times are applied on each layer in the simulation, $t_{j,i+1} - t_{j,i} = t_{layer} + t^{i}_{dwell}$, and $t_{j,i+2} - t_{j,i+1} = t_{layer} + t^{i+1}_{dwell}$, where $t_{layer}$ is the time to print one layer and $t^{i}_{dwell}$ is the dwell time for the $i$-th layer. In most cases, $t^{i+1}_{dwell}$ is larger than $t^{i}_{dwell}$, as more time is required for the higher layer to cool down considering the cooling rate reduces with the height [2].

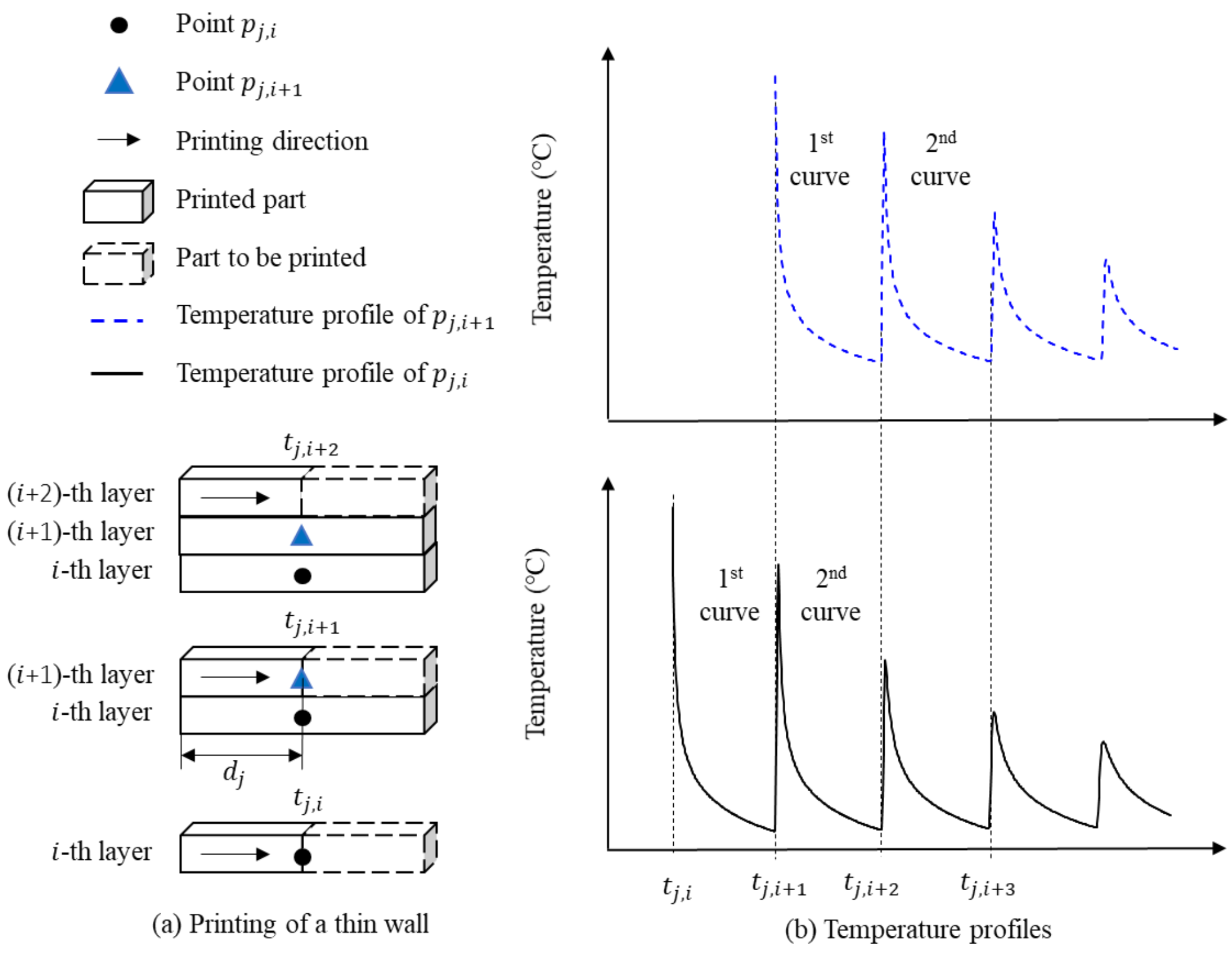

Figure 2 Similarity of temperature profiles between layers

Meanwhile, the curve of the point $p_{j,i+1}$ during $[t_{j,i+2}, t_{j,i+3}]$ seems similar to the curve of the point $p_{j,i}$ during $[t_{j,i+1}, t_{j,i+2}]$, as shown in the second curves in Figure 2 (b). The reason is that the point $p_{j,i}$ on the $i$-th layer is re-heated by the heat input on the ($i$+1)-th layer within $[t_{j,i+1}, t_{j,i+2}]$, and the point $p_{j,i+1}$ on the ($i$+1)-th layer is re-heated by the heat input on the ($i$+2)-th layer within $[t_{j,i+1}, t_{j,i+2}]$. For both points, the layer distance between the point layer and the heat-input layer is identical. This phenomenon is also applicable to any other two curves with the same index number.

Therefore, given any point pair $(p_{j,i}, p_{j,i+1})$, the $k$-th curve $\boldsymbol{C}^{k}_{j,i}$ in the temperature profile of $p_{j,i}$ is always similar to the $k$-th curve $\boldsymbol{C}^{k}_{j,i+1}$ in the temperature profile of $p_{j,i+1}$ considering the similar re-heating-cooling cycle encountered by both points. For clarification, this similarity between two successive layers is defined as "*curve similarity*", and $(\boldsymbol{C}^{k}_{j,i}, \boldsymbol{C}^{k}_{j,i+1})$ is defined as a "*curve pair*" containing such similarity.

## 2.3 Summary remarks

Both analyses are based on a thin wall with the same process setting on each layer, which means the printing direction, the travel speed, the wire feed rate, and other process parameters are fixed during the printing. Therefore, the time to print one layer is fixed for all layers. And the duration from depositing the material to re-heating the same point is identical for any points on the same layer. Although similarities in such a thin wall could be observed easily in the simulated temperature profiles, the thermal behavior in actual experiments is more complex considering the uncertainties in various factors, such as the printing process, material property, and geometry deviations. In this paper, several assumptions are applied as follows to simplify the effects of differences between simulations and actual experiments on modelling processes.

- *Constant Power*: In WAAM experiments, the metal wire is heated by the electric arc whose current $I$ and voltage $V$ change during the printing process. Although the dynamic update of current and voltage happens internally in the algorithm of the CMT Fronius apparatus, they are not captured in the FEA simulations. Therefore, in one single simulation or experiment, the power $P = I \cdot V$ is assumed as constant during printing.
- *Fixed Layer Thickness*: The layer thickness is set as a constant in FEA simulations, but the value varies in actual WAAM experiments considering the effects of geometry deviations. Specifically, although a constant step height is applied in the experiment, the thicknesses of printed layers are different and not equal to the step height. For simplification, the layer thickness in one experiment is assumed to be the same as the step height during the online prediction.
- *No Effect of Geometry Deviation*: The geometry deviation could affect the thermal behavior significantly during actual experiments, but they are only measured after the part is completed. In other words, the geometry deviation could not be an input variable for the online control. Therefore, the geometry deviation is assumed to not affect the profile similarities and curve similarities.

# 3 Methods

## 3.1 Data Preparation

Several simulations and WAAM experiments are completed for thin walls with unidirectional printing to collect temperature profiles by corresponding preprocessing methods for further discussion and testing.

### 3.1.1 Experimental setup

Based on the same machine system in the work [2], the experimental thermal data are collected from cold metal transfer based WAAM (CMT-WAAM) experiments, whose detailed setup is discussed in our previous work [14]. The travel speed $TS$, the wire feed rate $WFR$, and the shielding gas flow rate $SGFR$ are chosen as the process parameters to design several experiments, as they are the most effective parameters in the CMT-WAAM process [15]. Each parameter has three levels, i.e., $TS \in \{8,11,15\}$, $WFR \in \{3,4.5,6\}$, and $SGFR \in \{12,16,20\}$, to cover the $WFR/TS$ range [3.33,12.5] in the study [2] and the $SGFR$ values used in other CMT-WAAM studies [16–18]. In this paper, the central composite faced design of experiment is applied to generate 15 experiments, whose details are shown in Table 1.

In each experiment, a thin wall with a fixed length $160\ mm$ is printed with the AM70 alloy steel wire, whose diameter is $1.2\ mm$. Then the deposition rate, $DR$, is calculated as $3.14 \times 0.6^2 \times WFR \times 1000/60$. The baseplate is an S355 mild steel plate with size $300\ mm \times 50\ mm \times 20\ mm$. The step height $SH$ is set manually to make sure a successful printing of each wall, based on which the relative height of the $i$-th layer is estimated as $h_i = i \times SH$. The time to print one layer $t_{layer}$ is extracted from the ABB robot event log files. All these values are shown in Table 1.

Table 1 Details of 15 CMT-WAAM experiments

| Experiment No. | $TS$ $(mm/s)$ | $WFR$ $(m/min)$ | $SGFR$ $(L/min)$ | $SH$ (mm) | $t_{layer}$ (s) | $DR$ $(mm^3/s)$ |
|---|---|---|---|---|---|---|
| 1 | 8 | 3 | 12 | 1.6 | 20.2 | 56.52 |
| 2 | 8 | 6 | 20 | 1.6 | 20.2 | 113.04 |
| 3 | 8 | 6 | 12 | 2 | 20.2 | 113.04 |
| 4 | 8 | 3 | 20 | 2 | 20.2 | 56.52 |
| 5 | 15 | 3 | 20 | 1.4 | 10.9 | 56.52 |
| 6 | 15 | 3 | 12 | 1.4 | 10.9 | 56.52 |
| 7 | 15 | 6 | 12 | 1.6 | 10.9 | 113.04 |
| 8 | 15 | 6 | 20 | 1.6 | 10.9 | 113.04 |
| 9 | 11 | 4.5 | 16 | 1.5 | 14.8 | 84.78 |
| 10 | 8 | 4.5 | 16 | 1.8 | 20.2 | 84.78 |
| 11 | 15 | 4.5 | 16 | 1.5 | 10.9 | 84.78 |
| 12 | 11 | 3 | 16 | 1.5 | 14.8 | 56.52 |
| 13 | 11 | 6 | 16 | 1.8 | 14.8 | 113.04 |
| 14 | 11 | 4.5 | 12 | 1.6 | 14.8 | 84.78 |
| 15 | 11 | 4.5 | 20 | 1.7 | 14.8 | 84.78 |

During the printing, the dwell time is applied to make sure the interpass temperature is around 200 °C. The dwell time $t_{dwell}^{i}$ of the $i$-th layer is estimated as the difference between the absolute time from starting printing the $i$-th layer to starting printing the ($i$+1)-th layer and the calculated printing time $t_{layer}$. Due to the limited equipment, the temperature of the middle point on the side of the layer is detected every five layers. More specifically, the pyrometer is set to measure the temperature of the middle point on the first layer when printing the first layer to the fifth layer, to guarantee five temperature curves in each profile. When starting to print the sixth layer, the pyrometer is moved to record the temperature of the middle point on the sixth layer before printing the eleventh layer. Therefore, the point pair in experiments is defined as the middle points on the $i$-th layer and ($i$+5)-th layer, such as the first layer and the sixth layer. Considering the measurement capability of the applied pyrometer, the temperature above 1000 °C and the temperature below 150 °C would be truncated to the corresponding boundary as shown in Figure 3, which cannot reflect the actual thermal profile. In this paper, only experimental curves below 1000 °C are considered.

Different from the profile preprocessing for simulations in Section 3.1.2, the experimental temperature profile is split at each temperature sharp rise on the profile, shown as the red dashed line in Figure 3. The reason is that the exact locations of measured points are not accessible during experiments, which is attributed to the deviations caused by moving the pyrometer manually. In theory, such a split method has minor effects on testing the proposed prediction method. The key point of the work in this paper is to learn the curve similarity and the profile similarity, while these similarities are also observed among the obtained experimental temperature curves. Finally, 29 point pairs and 145 curve pairs are obtained from the experiments.

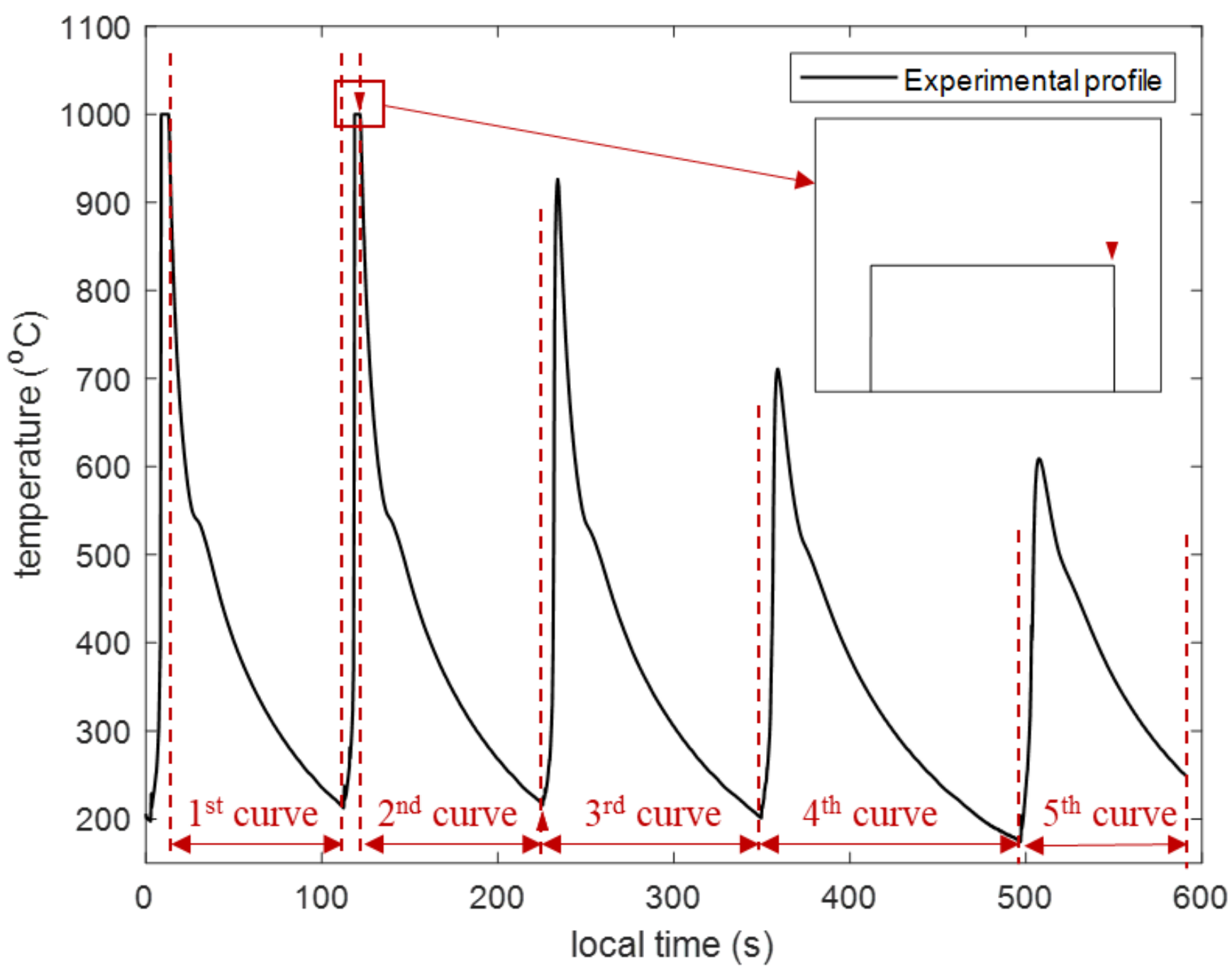

Figure 3 Temperature profile recorded by pyrometer

### 3.1.2 Simulation setup

The COMSOL Multiphysics software is applied to simulate the CMT-WAAM process. As the shielding gas is not applicable in the software, the travel speed $TS$ and the wire feed $WFR$ from completed experiments are selected to design nine simulations, as shown in Table 2. The materials in simulations are identical to experiments, i.e., S355 mild steel for the baseplate with a fixed geometry size $300\ mm \times 200\ mm \times 20\ mm$, and AM70 alloy steel for the part.

In each simulation, a thin wall with a layer width $4.4\ mm$, a layer length $160\ mm$, and 40 layers is studied. The layer thickness $l_t$ is set according to step heights applied in actual WAAM experiments. Then the relative height of the $i$-th layer is calculated as $h_i = i \times l_t$. And the deposition rate $DR$ is estimated as $4.4 \times l_t \times TS$ for the simulation. The printing time $t_{layer}$ for one layer is set as $160/TS + 0.5$, where $0.5$ is a constant total estimation of the acceleration time at the starting point and the deceleration time at the ending point of one layer.

Table 2 Details of COMSOL Multiphysics simulations

| Simulation No. | $TS$ $(mm/s)$ | $WFR$ $(m/min)$ | $t_{layer}$ $(s)$ | $DR$ $(mm^3/s)$ | $l_t$ $(mm)$ |
|---|---|---|---|---|---|
| 1 | 8 | 3 | 20.5 | 52.8 | 1.5 |
| 2 | 8 | 6 | 20.5 | 70.4 | 2.0 |
| 3 | 15 | 3 | 11.17 | 92.4 | 1.4 |
| 4 | 15 | 6 | 11.17 | 105.6 | 1.6 |
| 5 | 11 | 4.5 | 15.05 | 77.44 | 1.6 |
| 6 | 8 | 4.5 | 20.5 | 63.36 | 1.8 |
| 7 | 15 | 4.5 | 11.17 | 99 | 1.5 |
| 8 | 11 | 3 | 15.05 | 72.6 | 1.5 |
| 9 | 11 | 6 | 15.05 | 87.12 | 1.8 |

To collect sufficient temperature profiles, seven points are set evenly on the side of each layer, as shown in Figure 4. According to the printing direction, the point $p_{j,i}, j \in [1,7]$ is located with a distance $20j\ mm$ to the starting point of the $i$-th layer. Therefore, seven point pairs could be generated among every two successive layers in the simulation.

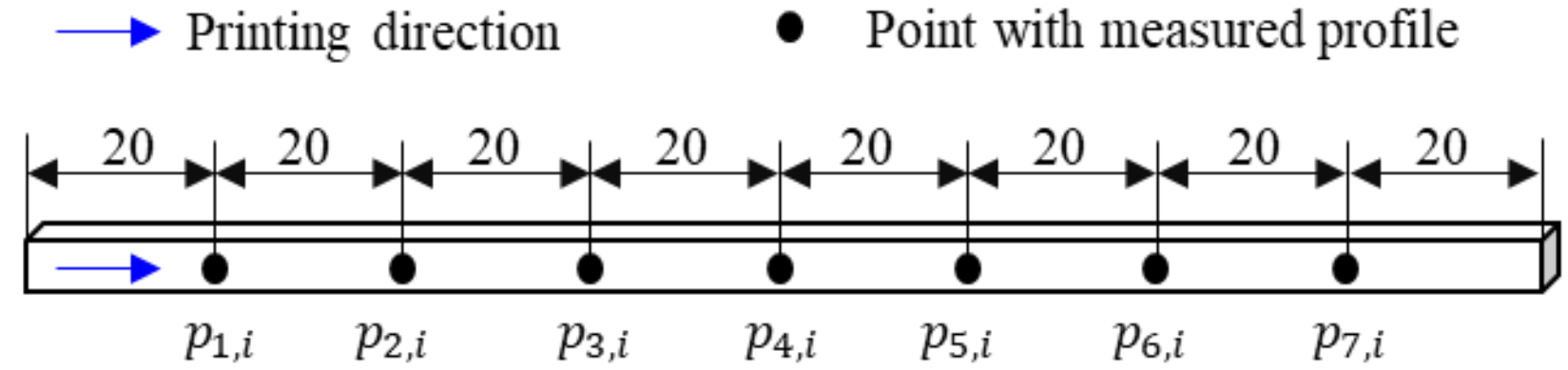

Figure 4 Points located on the $i$-th layer in simulations

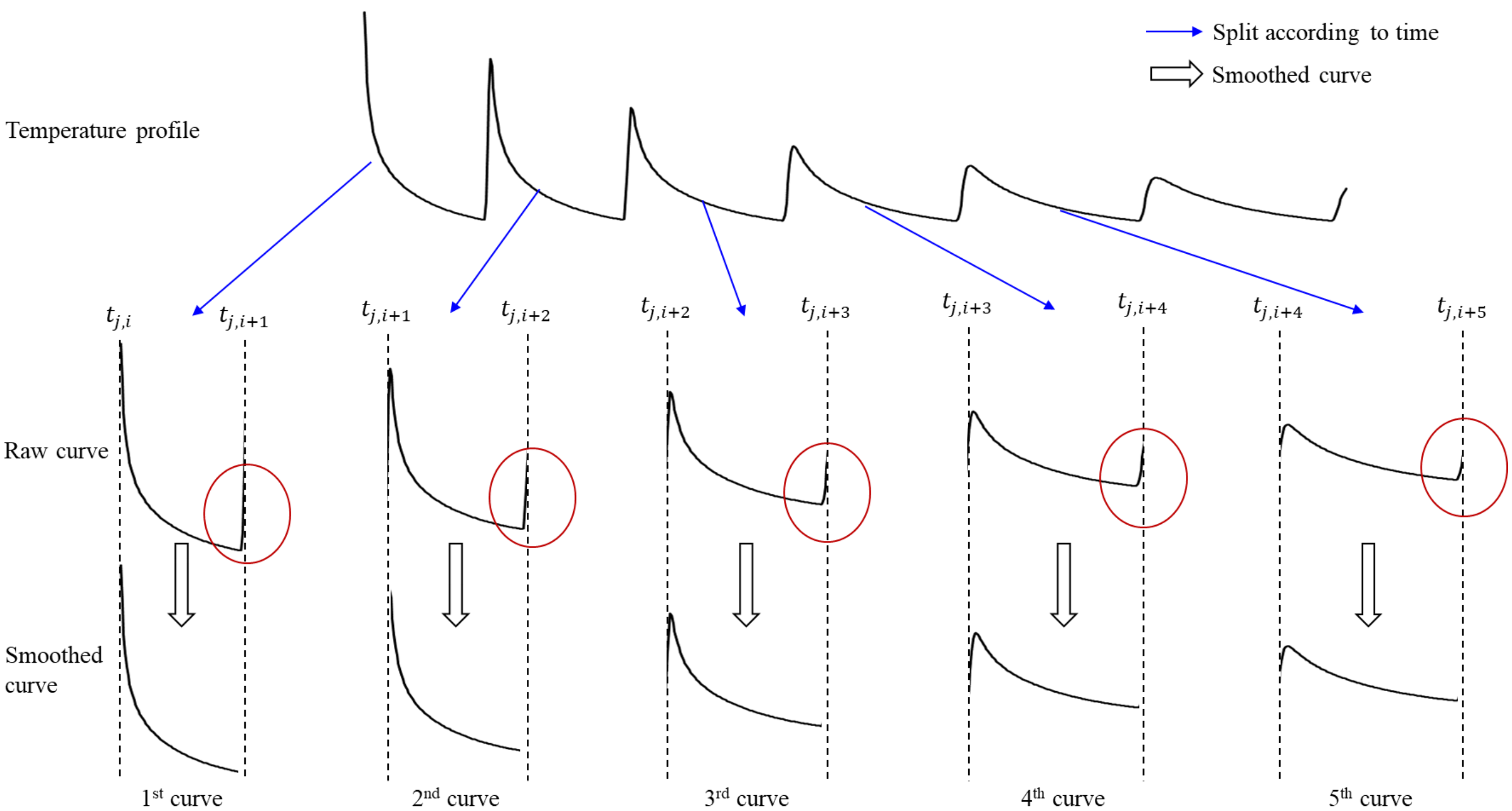

Figure 5 Temperature profile vs temperature curves

For the temperature profile of any point $p_{j,i}$ in the thin wall, it encounters multiple "*re-heating-cooling*" cycles according to the printing and cooling on higher layers, as shown in Figure 5. Given the travel speed $TS$, the printing time $t_{layer}$ of one layer, the dwell time $t_{dwell}^{i}$ of each $i$-th layer, and the distance $d_j$ between $p_{j,i}$ and the boundary point $p_{0,i}$, the global time $t_{j,i}$ when the material is deposited to $p_{j,i}$ could be estimated offline or online as follows.

$$t_{j,i} = (i-1)t_{layer} + \sum_{m=1}^{i-1} t_{dwell}^{m} + d_j/TS \tag{1}$$

Based on the thermal behavior discussed in Section 2, the temperature profile could be split into several curves according to the estimated time series $\{t_{j,i}, t_{j,i+1}, t_{j,i+2}, \ldots\}$ as shown in Figure 5. For clarification, the partial profile within $[t_{j,i+k-1}, t_{j,i+k}]$ is defined as the $k$-th temperature curve of the point $p_{j,i}$, whose duration is a fixed value for all points on the same layer, i.e., $t_{j,i+k} - t_{j,i+k-1} = t_{layer} + t_{dwell}^{i+k-1}$. This phenomenon has also been observed in Section 2.1. However, a short-term increase could be observed in the tail of each raw curve, as shown in the red circles in Figure 5. This reason is that the element at one point would be re-heated when the distance between the heat input on the higher layer and the point is smaller than a certain value in the FEA simulation, as well as the effect of heat transfer. For simplification, this short-term increase will not be considered in the online prediction method in Section 3.2, as its duration is negligible compared to the entire duration of one curve. Therefore, the smoothed curves after processing are selected for further application in the paper.

Besides, the durations of all temperature curves at one point are different, as the dwell time increases with the layer, i.e., $t_{dwell}^{i+1} \geq t_{dwell}^{i}$. To represent these temperature curves with various durations, a fixed number $N$ of discrete temperature data is sampled evenly from each curve. Therefore, the $k$-th temperature curve $\boldsymbol{C}_{j,i}^{k}$ of point $p_{j,i}$ could be formulated as $\boldsymbol{C}_{j,i}^{k} = [T_{j,i}^{k,1}, \dots T_{j,i}^{k,n}, \dots, T_{j,i}^{k,N}]$, where $T_{j,i}^{k,n}$ is the $n$-th sampled temperature data.

Theoretically, as the remaining number of layers decreases as the geometry is built, the number of re-heating cycles is reduced. Hence, the number of temperature curves decreases with the increase of layers. In this paper, for consistency only the first five curves, also the most influential curves, on the temperature profile of every point are selected for discussion. As a result, the temperature profile $\boldsymbol{P}_{j,i}$ of point $p_{j,i}$ is represented as $\boldsymbol{P}_{j,i} = [\boldsymbol{C}_{j,i}^{1}, \boldsymbol{C}_{j,i}^{2}, \boldsymbol{C}_{j,i}^{3}, \boldsymbol{C}_{j,i}^{4}, \boldsymbol{C}_{j,i}^{5}]$.

Based on the above profile processing, the collected temperature profiles from one simulation can generate 1,260 curve pairs in total, which means 11,340 curve pairs from all nine simulations.

### 3.2 Online Two-stage Prediction Method

In actual WAAM experiments, several points on a layer would be selected to measure the thermal history instead of the entire part due to physical and practical limitations. Consequently, the thermal history of most points in the part would be lost, which poses a challenge to applying thermal information for *in-situ* control. To alleviate the problem, an online two-stage thermal history prediction method is proposed and designed based on the "*curve similarity*" between two successive layers and the "*profile similarity*" within one layer. A high-level description of the method is shown in Figure 6 and described as follows. More details are discussed in Sections 3.2.1 and 3.2.2.

- *Stage 1: Layer-to-layer prediction*. This stage aims to train a prediction model to learn the curve similarity between the $k$-th temperature curves of any point pair $(p_{j,i}, p_{j,i+1})$. If the thermal history could be measured online during printing, the temperature curves of points on the yet-to-print layer could be predicted from the curves of counterpart points on the lower printed layer by the trained model.
- *Stage 2: Intra-layer prediction*. The prediction model in this stage is designed to learn the profile similarity within one layer. Based on the predicted or measured temperature profiles of several points on one layer, the constructed intra-layer prediction model could estimate the temperature profiles of unseen points on the same layer, which in turn forms the temperature field of the entire layer.

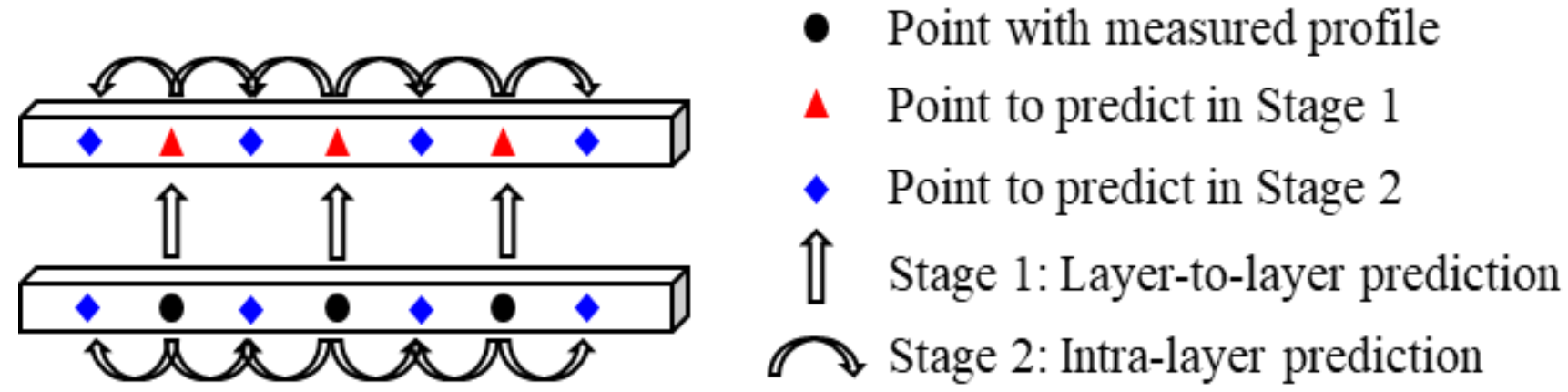

Figure 6 A high-level description of the online two-stage prediction method

#### 3.2.1 Stage 1: Layer-to-layer prediction

**(1) Modeling objective**

Based on the discussion in Section 2.2, given the $k$-th temperature curve $\boldsymbol{C}_{j,i}^{k}$ of the point $p_{j,i}$ on the $i$-th layer, the counterpart one with similarity is the $k$-th temperature curve $\boldsymbol{C}_{j,i+1}^{k}$ of the point $p_{j,i+1}$ on the ($i$+1)- th layer. If the curve similarity is formulated as a model $f_{s1}(\cdot)$, the curve $\boldsymbol{C}_{j,i+1}^{k}$ could be represented as a function $\boldsymbol{C}_{j,i+1}^{k} = f_{s1}(\boldsymbol{C}_{j,i}^{k}, \boldsymbol{x})$, where $\boldsymbol{x}$ contains other input variables. However, as $dt_i \leq dt_{i+1}$, the duration of the curve $\boldsymbol{C}_{j,i}^{k}$, i.e., $t_{layer} + t_{dwell}^{i+k-1}$, is smaller than the duration of the curve $\boldsymbol{C}_{j,i+1}^{k}$, i.e., $t_{layer} + t_{dwell}^{i+k}$. In other words, the length of the curve $\boldsymbol{C}_{j,i}^{k}$ is shorter than the curve $\boldsymbol{C}_{j,i+1}^{k}$ as shown in Figure 7, where the starting time of each curve is defined as the local time 0. To maintain the curve similarity when training $f_{s1}(\cdot)$, only the partial curve $\boldsymbol{C}_{j,i+1}^{k'}$ within the overlapped duration on $\boldsymbol{C}_{j,i+1}^{k}$ is defined as the output of the model, i.e., $\boldsymbol{C}_{j,i+1}^{k'} = f_{s1}(\boldsymbol{C}_{j,i}^{k}, \boldsymbol{x})$. In this paper, both $\boldsymbol{C}_{j,i+1}^{k'}$ and $\boldsymbol{C}_{j,i+1}^{k}$ are represented as an $1 \times N$ vector with different sampling periods as discussed in Section 3.1.2.

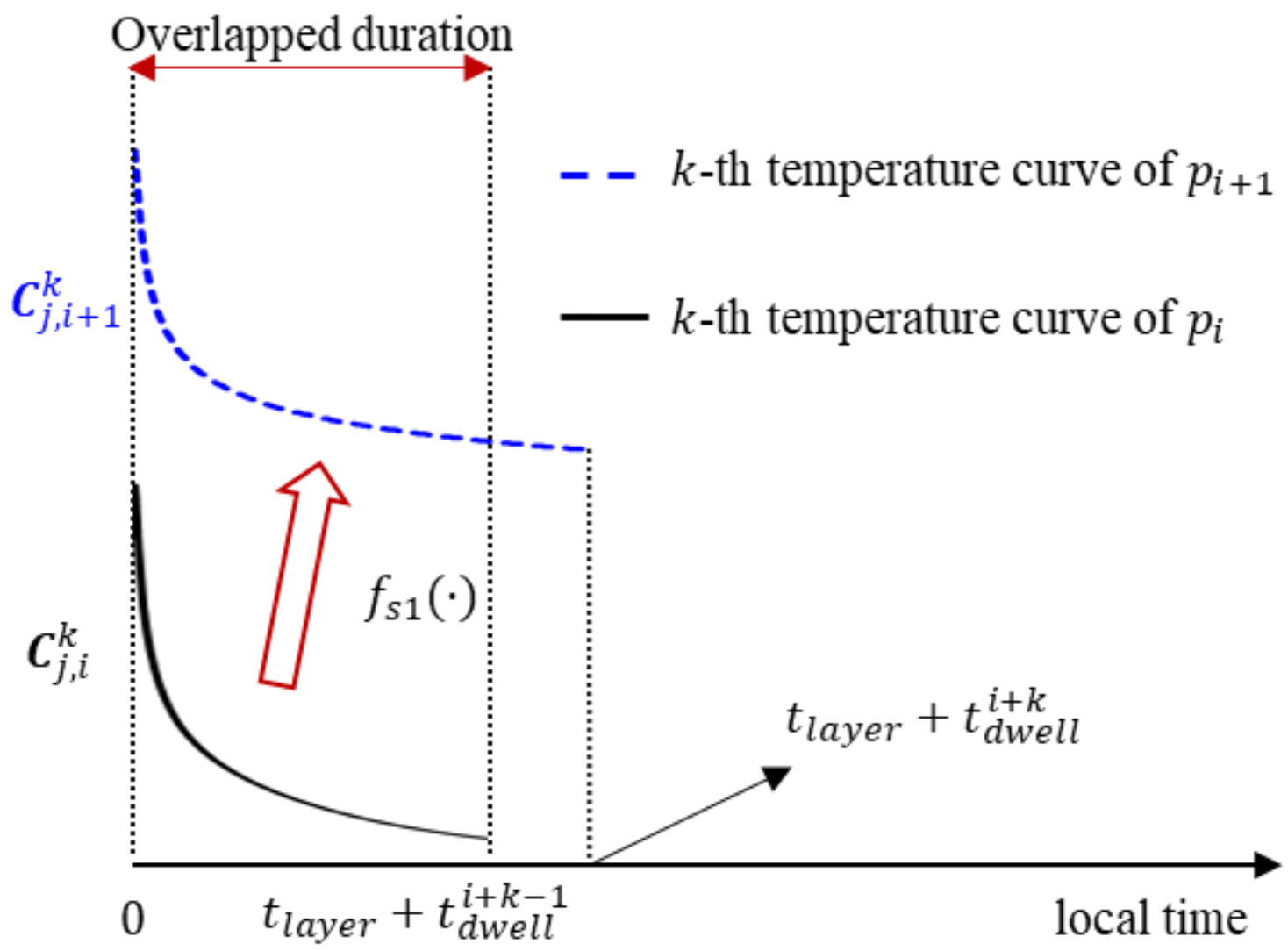

Figure 7 Illustration of $k$-th curves of a point pair under local time

According to the study [13], each temperature curve follows an exponential formulation, which is attributed to the exponential re-heating and cooling behavior. This indicates that the predicted partial curve $\boldsymbol{C}_{j,i+1}^{k'}$ could be used to extrapolate the entire curve $\boldsymbol{C}_{j,i+1}^{k}$ based on the local time, which provides a possibility to predict the dwell time of the yet-to-print layer to cool down to the required interpass temperature. Finding the global optimal modeling method, however, is not the purpose of the paper, the extrapolation from $\boldsymbol{C}_{j,i+1}^{k'}$ to $\boldsymbol{C}_{j,i+1}^{k}$ will not be considered in the following sections for simplification. Therefore, the profile of the point on the yet-to-print layer is only represented with partial curves in this work, i.e., $\boldsymbol{P}_{j,i} = [\boldsymbol{C}_{j,i}^{1'}, \boldsymbol{C}_{j,i}^{2'}, \boldsymbol{C}_{j,i}^{3'}, \boldsymbol{C}_{j,i}^{4'}, \boldsymbol{C}_{j,i}^{5'}]$.

**(2) Input variables**

To improve the generalization of the model $f_{s1}(\cdot)$ in various applications (e.g., online/offline prediction of thin walls with different process settings), several parameters are considered complementary variables, i.e., $\boldsymbol{x} = [t_{layer}, t_{dwell}^{i}, DR, h_i]$.

- *Printing time of one layer*, $t_{layer}$ (*unit*: $s$): The heat input in WAAM experiments is the energy per unit length, which is formulated as $P/TS$ with the unit $J/mm$ [19]. Given the length of one layer $L$ ($mm$), the total heat input on the layer is formulated as $P \cdot L/TS$. As the power $P$ is assumed as a constant in this paper and the modeling purpose of $f_{s1}(\cdot)$ is to learn the curve similarity, the effect of heat input on the curve could be simplified and reflected by the time to print one layer, i.e., $t_{layer} = L/TS$.
- *Dwell time of the* $i$*-th layer*, $t_{dwell}^{i}$ (*unit*: $s$): When one layer in the thin wall is printed completely, the dwell time is taken until the interpass temperature cools down to the required value. Although the cooling rate during this process depends on several factors (e.g., the time, the radiation/convection coefficient, and the external cooling system), only the dwell time is selected as the variable to indicate the effect of cooling rate on the temperature curve in this paper.

- *Deposition rate*, $DR$ (*unit*: $mm^3/s$): In actual WAAM experiments, the deposition rate could be estimated as $DR = \pi \times (d/2)^2 \times WFR \times 1000/60$, where $d$ $(mm)$ is the diameter of the metal wire, and $WFR$ $(m/min)$ is the wire feed rate in the process. Therefore, the volume of one printed layer is estimated as the product of the printing time and the deposition rate. Moreover, as a larger $WFR$ value indicates a higher input [20], the deposition rate could be a general variable to reflect the total heat input in one layer when metal wires of various diameters are used.
- *Relative height of the* $i$*-th layer*, $h_i$ (*unit*: $mm$): In WAAM processes, the difference in the initial thermal conditions to print each layer decreases with the height. More specifically, the initial thermal condition to print the first layer is related to the substrate only, while the initial condition for the second layer depends on the thermal field of the first layer. As a certain interpass temperature is maintained during printing, the difference between the initial thermal conditions of the first and the second layers would be larger than the differences between the second and the third layer. One reflection of the difference reduction is that the cooling rate would decrease gradually to a stable value with height. Another reflection is the increasing curve similarity, which means the difference between the $k$-th curves of a point pair $(p_{j,i}, p_{j,i+1})$ decreases with height, as shown in Figure 8. The input curve refers to the temperature curve of the point on the lower layer, while the output one is the curve of the point on the higher layer. To reflect the effect of the height in the model $f_{s1}(\cdot)$, the relative height $h_i$ of the $i$-th layer to the substrate is defined as one input variable, i.e., $h_i = i \cdot l_t$, where $l_t$ is the constant layer thickness.

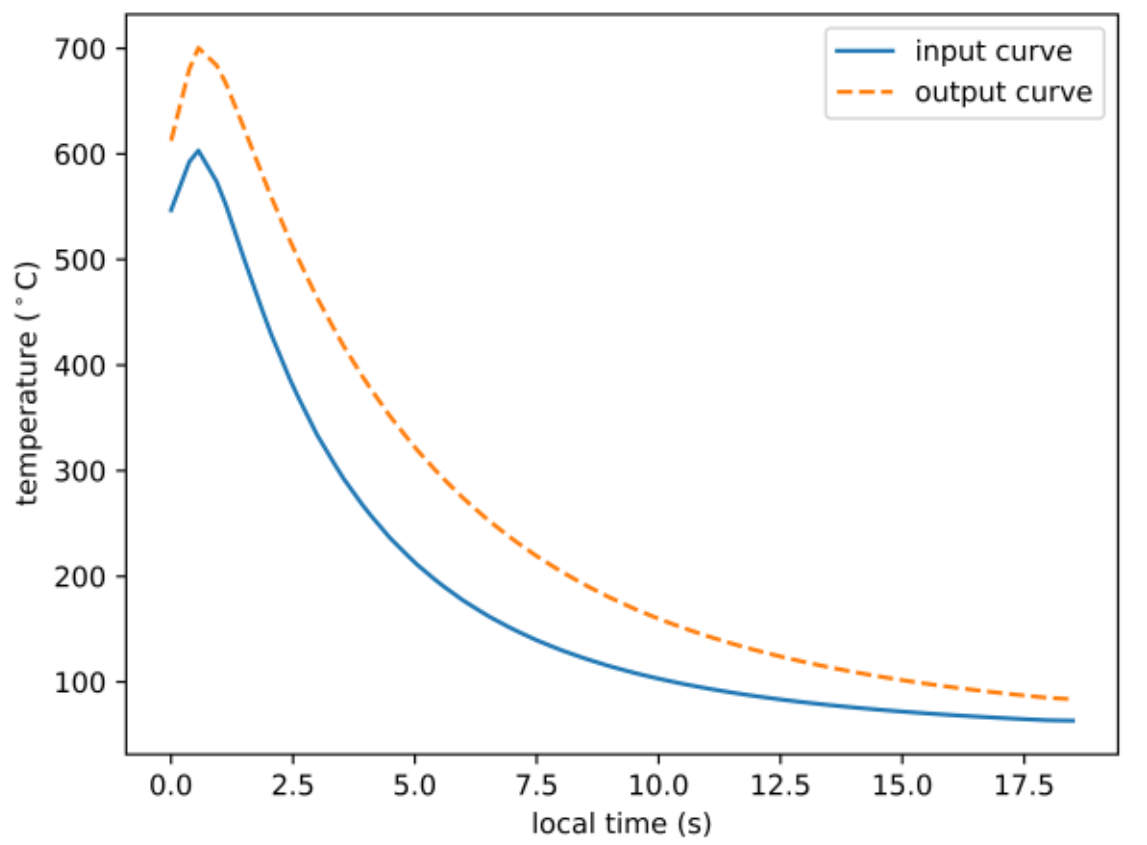

(a) Curve of the point pair on the 1st and 2nd layer

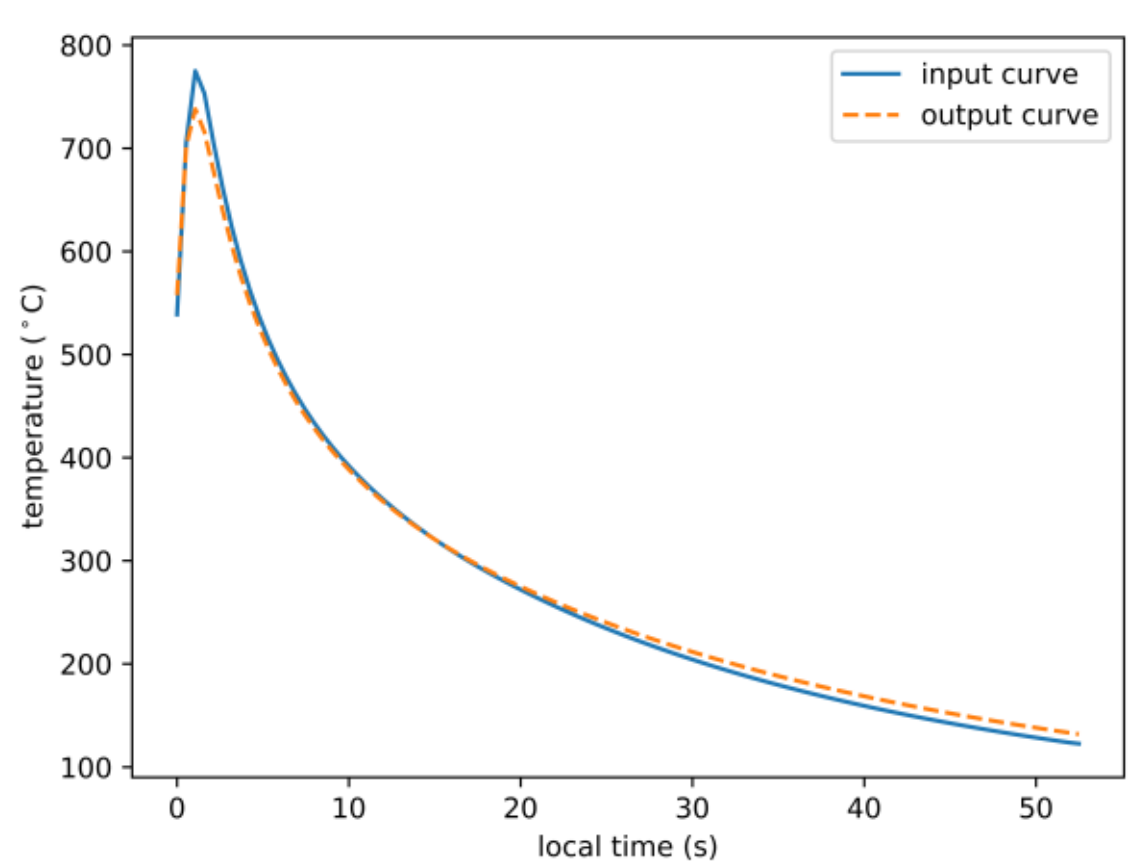

(b) Curve of the point pair on the 10th and 11th layer

Figure 8 Similarity between the third curves of point pairs

**(3) Model structure**

Based on the above discussions, the layer-to-layer prediction model is finally formulated as $\boldsymbol{C}_{j,i+1}^{k'} = f_{s1}(\boldsymbol{C}_{j,i}^{k}, t_{layer}, t_{dwell}^{i}, DR, h_i)$, which has $N+4$ input variables and $N$ outputs. This means the proposed model in the first stage is a multi-input-multi-output (MIMO) regression task. In this paper, the artificial neural network is chosen to construct the model $f_{s1}(\cdot)$, considering its ability to process noisy data (i.e., the experimental temperature data), easy implementation, and vast successful applications in different MIMO tasks, such as nonlinear control systems [21], image segmentation [22], and transportation [23].

Theoretically, an optimal neural network exists for the layer-to-layer prediction model $f_{s1}(\cdot)$ to receive the best performance. Generally, an optimization problem is solved to find optimal hyperparameters of the neural network, including the number of hidden layers, the number of hidden neurons on each layer, the activating function for each hidden layer, the learning rate, the epoch number for training, and other model structure parameters. This is the topic of "*hyperparameter optimization*" in deep learning [24], which is not covered in this paper, however.

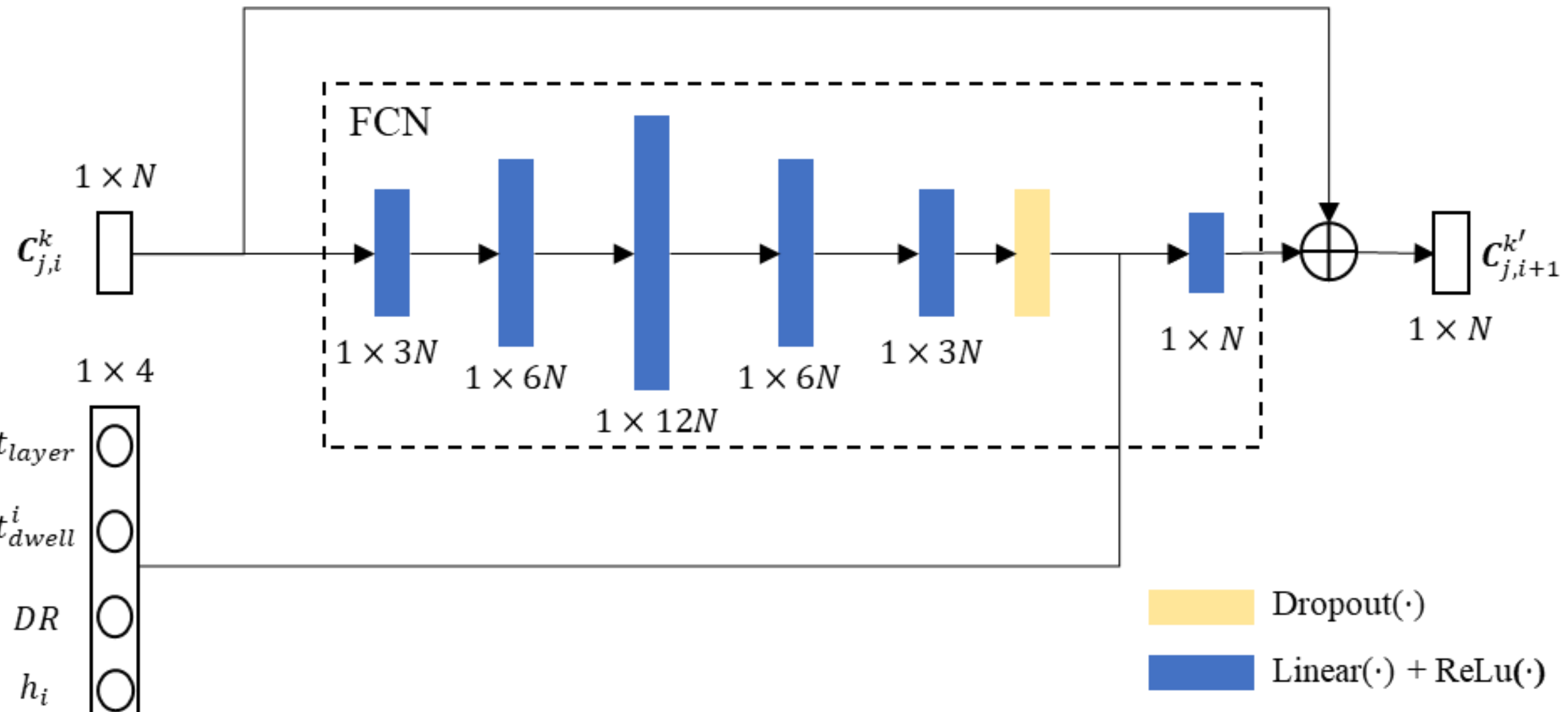

Figure 9 Structure of the layer-to-layer prediction model $f_{s1}(\cdot)$

Instead of solving a hyperparameter optimization problem, a fully connected network with residual connection (FCN-RC) is selected as $f_{s1}(\cdot)$ after several trials, as it provides acceptable testing performance for this problem. The model structure is shown in Figure 9. A linear connection layer Linear(·) and the activation function ReLu(·) are chosen for each hidden layer. As the size of elements in $\boldsymbol{C}_{j,i}^{k}$ could be set as various values, the number of hidden nodes on each layer is defined according to $N$, i.e., $3N$, $6N$, $12N$, $6N$, $3N$, $N$ nodes are set from the first to the sixth hidden layer respectively. In theory, there are total $186N^2 + 35N$ parameters for learning in the model. A dropout layer with a rate 0.1 is applied before the output layer to reduce the risk of overfitting. Then a residual connection is used to combine the input curve $\boldsymbol{C}_{j,i}^{k}$ to form the final output, i.e., $\boldsymbol{C}_{j,i+1}^{k'} = FCN\left(\boldsymbol{C}_{j,i}^{k}, t_{layer}, t_{dwell}^{i}, DR, h_i\right) + \boldsymbol{C}_{j,i}^{k}$. Given the number of training epoch, the residual connection could make the training easier and improve learning performance [25]. During training, the Adam optimizer [26] is applied with the mean square error loss, with the initial learning rate 0.001 and the total training epoch number 500. Besides, the learning rate is reduced by a ratio of 0.5 after every 100 epochs to improve the convergence. In this paper, all codes are implemented based on Pytorch [27].

### 3.2.2 Stage 2: Intra-layer prediction

#### (1) Modeling objective

In this stage, only temperature profiles of points on the same layer are considered. For instance, discrete points are located on the $i$-th layer, denoted as $p_{j,i}, j \in [1, M]$. The temperature profile $\boldsymbol{P}_{j,i}$ of the point $p_{j,i}$ is measured by sensors or constructed from the temperature curves predicted by $f_{s1}(\cdot)$. As mentioned in Section 2.1, the only factor to differentiate each point is their distance $d_j$ to the boundary point of the layer or the relative delay $t_{rd}^{j} = d_j/TS$. Considering the observed profile similarity, the relationship between the relative delay and the corresponding temperature profile could be modeled as a function, i.e., $\boldsymbol{P}_{j,i} = f_{s2}(t_{rd}^{j})$, where $f_{s2}(\cdot)$ is the intra-layer prediction model. Then, given an unseen point $p'$ whose distance to the boundary is $d'$ and relative delay is $t_{rd}' = d'/TS$, its temperature profile $\boldsymbol{P}'$ could be estimated as $\boldsymbol{P}' = f_{s2}(t_{rd}')$ directly.

**(2) Model structure**

According to Section 3.1, each profile is constructed with five predicted curves, whose dimension is then $1 \times 5N$. Therefore, the training of the intra-layer prediction model $f_{s2}(\cdot)$ is a single-input-multi-output regression task. To keep the accuracy of the curve representation $\boldsymbol{C}_{j,i}^{k}$, the size $N$ of temperature data sampled on each curve is set as a large value, then the size $5N$ would be in hundreds or thousands. However, considering the limitation of sensors and the constraints of geometry size, $M$ is in tens (also the training data size), which means $5N$ is much larger than $M$ in actual experiments. In this case, the training performance of the high-dimension-output model would be poor; and the overfitting problem would exist and the generalization performance is worse on testing data.

To decrease the huge diversity between the training data size and the output size, the nonintrusive reduced-order modeling (ROM) method is applied for dimension reduction in this paper. It is a kind of technology to decompose the responses of complex systems with a high degree of freedom into low-dimensional intrinsic features [28]. For example, given a bunch of responses (snapshots) from the same system with various boundary conditions, multiple common features could be extracted from the response matrix by model decomposition methods, such as the proper orthogonal decomposition (POD) [29]. Then each snapshot could be represented as a combination of all common features, where the feature coefficients are modeled as a function of the boundary conditions. When the raw snapshots are decomposed directly without any nonlinear preprocessing, the obtained common features are linear combinations of snapshots with various coefficients. Therefore, the final ROM solution for a new boundary condition is also a linear combination of snapshots. Generally, the accuracy of the ROM solution increases with the linearity observed [29]. Due to the efficiency and accuracy of ROM methods in constructing and representing high-dimensional responses, the non-intrusive ROM method has been applied widely in various engineering problems, such as heat transfer for unsteady flow [30], natural convection in porous media [31], and structure responses [32].

*(a) Profile decomposition by POD*

Based on the profiles $\boldsymbol{P}_{j,i}$ of all point $p_{j,i}$ ($j \in [1, M]$) on the $i$-th layer, a profile matrix $\boldsymbol{S}$ is defined as $\boldsymbol{S} = [\boldsymbol{P}_{1,i}^{T}, \dots, \boldsymbol{P}_{j,i}^{T}, \dots, \boldsymbol{P}_{M,i}^{T}]$ with the size $5N \times M$, where $5N \gg M$, and $\boldsymbol{P}_{j,i}^{T}$ is the transposition of the profile vector $\boldsymbol{P}_{j,i}$. Then the singular value decomposition of the profile matrix is conducted as:

$$\boldsymbol{S} = \boldsymbol{U\Sigma V}^{T} = [\boldsymbol{u}_1, \dots, \boldsymbol{u}_{5N}]_{5N\times 5N} \begin{bmatrix} \lambda_1 & 0 & 0 \\ 0 & \dots & 0 \\ 0 & 0 & \lambda_M \\ & \boldsymbol{0} & \end{bmatrix}_{5N\times M} [\boldsymbol{v}_1, \dots, \boldsymbol{v}_M]_{M\times M} \tag{2}$$

where $\boldsymbol{u}_1, \dots, \boldsymbol{u}_{5N}$ are the left singular vectors, $\boldsymbol{v}_1, \dots, \boldsymbol{v}_M$ are the right singular vectors, and $\lambda_1, \dots., \lambda_M$ are the singular values with the relationship $\lambda_1 \geq \lambda_2 \geq \cdots \geq \lambda_M \geq 0$. As $\boldsymbol{0}$ is a zero matrix with size $(5N - M) \times M$, the above equation could be simplified as follow.

$$\boldsymbol{S} = [\boldsymbol{u}_1, \dots, \boldsymbol{u}_M]_{5N\times M} \begin{bmatrix} \lambda_1 & 0 & 0 \\ 0 & \dots & 0 \\ 0 & 0 & \lambda_M \end{bmatrix}_{M\times M} [\boldsymbol{v}_1, \dots, \boldsymbol{v}_M]_{M\times M} \tag{3}$$

A coefficient matrix is defined as $\boldsymbol{C} = diag(\lambda_1, \dots, \lambda_M)[\boldsymbol{v}_1, \dots, \boldsymbol{v}_M] = [\boldsymbol{C}_1, \dots, \boldsymbol{C}_M]$, where $\boldsymbol{C}_1, \dots, \boldsymbol{C}_M$ are the coefficient vectors with an identical size $M \times 1$. Then, the formulation in Eq. (3) is rewritten as:

$$\boldsymbol{S} = [\boldsymbol{u}_1, \dots, \boldsymbol{u}_M]_{5N\times M}[\boldsymbol{C}_1, \dots, \boldsymbol{C}_M]_{M\times M} = [\boldsymbol{P}_{1,i}^{T}, \dots, \boldsymbol{P}_{j,i}^{T}, \dots, \boldsymbol{P}_{M,i}^{T}] \tag{4}$$

Therefore, each temperature profile $\boldsymbol{P}_{j,i}^{T}$ could be calculated as:

$$\boldsymbol{P}_{j,i}^{T} = [\boldsymbol{u}_1, \dots, \boldsymbol{u}_M]_{5N\times M}\left[\boldsymbol{C}_j\right]_{M\times 1} = \sum_{m=\boldsymbol{1}}^{M} \boldsymbol{u}_{\boldsymbol{m}} c_j^m \tag{5}$$

In the field of ROM, $\boldsymbol{u}_1, \dots, \boldsymbol{u}_M$ are defined as the reduced bases, which are identical for each column in the matrix $\boldsymbol{S}$. To find the minimum number $M^*$ of the required reduced bases while maintaining the accuracy of the final solution, the energy percentage $\varepsilon$ defined based on the singular value is used as the criteria [33].

$$\varepsilon = \frac{\sum_{m=1}^{M^*} \lambda_m^2}{\sum_{m=1}^{M} \lambda_m^2} \geq 99\% \tag{6}$$

After obtaining the optimal number, the former $M^*$ reduced bases (left singular vectors) and their corresponding coefficients in Eq. (5) are selected to reformulate the $\boldsymbol{P}_{j,i}^T$ as the following equation:

$$\boldsymbol{P}_{j,i}^T \approx \widehat{\boldsymbol{U}}\widehat{\boldsymbol{C}}_j = [\boldsymbol{u}_1, \dots, \boldsymbol{u}_{M^*}]_{5N \times M^*} \left[c_j^1, \dots, c_j^{M^*}\right]^T = \sum_{m=1}^{M^*} \boldsymbol{u}_{\boldsymbol{m}} c_j^m \tag{7}$$

where $\widehat{\boldsymbol{U}} = [\boldsymbol{u}_1, \dots, \boldsymbol{u}_{M^*}]_{5N \times M^*}$ is the reduced basis matrix with selected bases, and $\widehat{\boldsymbol{C}}_j$ contains the basis coefficients of the point $p_{j,i}$.

*(b) Profile construction by ELM*

After decomposing the profile matrix $\boldsymbol{S}$ of the $i$-th layer, the profile $\boldsymbol{P}_{j,i}$ of the point $p_{j,i}$ only depends on the basis coefficients $\widehat{\boldsymbol{C}}_j$, while the reduced bases $\boldsymbol{u}_1, \dots, \boldsymbol{u}_{M^*}$ are identical for all points on the same layer. Therefore, the relationship between the relative delay $t_{rd}^j$ and the profile $\boldsymbol{P}_{j,i}$, i.e., $\boldsymbol{P}_{j,i} = f_{s2}(t_{rd}^j)$, could be simplified as the one between $t_{rd}^j$ and the basis coefficients, i.e., $\widehat{\boldsymbol{C}}_j = f_{s2'}(t_{rd}^j)$. In other words, the output dimension of the model to train reduces significantly from $5N$ to $M^*$ ($M^* \leq M \ll 5N$), which could make the training process easier for this stage.

The training data (basis coefficients), however, are only available after decomposing the temperature profiles of one layer, and such coefficients would vary significantly between layers. This indicates that the model $f_{s2'}(\cdot)$ cannot be transferred directly from one layer to another layer. Therefore, the model $f_{s2'}(\cdot)$ is constructed online for each layer, which poses a strict requirement for modeling efficiency.

In this paper, the extreme learning machine (ELM) is selected as the modeling method for $f_{s2'}(\cdot)$. ELM is a single-hidden layer feedforward neural network, which converges faster than the conventional feedforward neural network [34]. Given the training data $(\boldsymbol{X}, \boldsymbol{Y})$ and the size $N_h$ of hidden nodes, the general formulation of ELM is presented as:

$$f_{ELM}(\boldsymbol{X}) = \Sigma_{i=1}^{N_h} \beta_i g_i(\boldsymbol{\omega}_i \boldsymbol{X} + b_i) \tag{8}$$

where $\beta_i$ is the weight between the $i$-th hidden neuron and the output layer, and $g_i(\cdot)$ is the activation function. In ELM, only the weights $\boldsymbol{\beta} = [\beta_1, \dots, \beta_L]$ are learned, while the parameters ($\boldsymbol{\omega}_i$ and $b_i$) in activation functions are assigned randomly and frozen during training. Therefore, the optimization problem to train ELM is:

$$\min_{\beta_1, \dots, \beta_L} \|\boldsymbol{H}\boldsymbol{\beta} - \boldsymbol{Y}\|^2 \tag{9}$$

where $\boldsymbol{H}$ is a matrix containing the outputs of all hidden layers. Different from tuning both weights and activation function parameters in the iteration-based learning algorithm, ELM has much fewer parameters to learn and has no iteration, which expedites the training in general [35]. Moreover, the activation functions with random parameters could provide more statistical information about inputs and a universal approximation ability, which could provide a better generalization performance. It's proved that ELM could reach the optimal performance with a higher possibility than conventional networks [36]. Due to its outstanding performance, ELM has been applied widely in various real-time classification, clustering, and regression tasks, such as traffic flow prediction [37], transport estimation in pipes [38], and marketing price forecasting [39].

Given the relative delay and corresponding bases coefficients of all points on the layer, the input $\boldsymbol{X}$ and output $\boldsymbol{Y}$ to train the ELM model $f_{ELM}(\cdot)$ are defined as:

$$\boldsymbol{X} = \begin{bmatrix} t_{rd}^1 \\ \vdots \\ t_{rd}^M \end{bmatrix}, \boldsymbol{Y} = \begin{bmatrix} \widehat{\boldsymbol{C}}_1 \\ \vdots \\ \widehat{\boldsymbol{C}}_M \end{bmatrix} = \begin{bmatrix} c_1^1, \dots, c_1^{M^*} \\ \vdots \\ c_M^1, \dots, c_M^{M^*} \end{bmatrix} \tag{10}$$

After training, the basis coefficients $\widehat{\boldsymbol{C}}'$ of an unseen point $p'$ with a relative delay $t'_{rd}$ on the same layer could be estimated as:

$$\widehat{\boldsymbol{C}}' = f_{ELM}(t'_{rd}) \tag{11}$$

Finally, the temperature profile $\boldsymbol{P}'$ of the point $p'$ is constructed as the product of common features $\widehat{\boldsymbol{U}}$ and the estimated coefficients $\widehat{\boldsymbol{C}}'$.

$$\boldsymbol{P}' = \widehat{\boldsymbol{U}}\widehat{\boldsymbol{C}}' \tag{12}$$

In this paper, the open-source library $elm$ [40] is applied to implement the ELM model. The number of hidden nodes is set as 128, resulting in $128 \times (2 + M^*)$ parameters in the ELM model. The activation function is set as ReLu(·). And all other hyperparameters are set as the default values. For clarification, the combination of profile decomposition with POD and construction with ELM is defined as the intra-layer prediction model $f_{s2}(\cdot)$ in further discussions.

### 3.3 Summary remarks

Based on the above discussions, the steps of online prediction for a yet-to-print layer are described as follows.

- *Step 1*: On the $i$-th layer, temperature profiles of $M$ discrete points $p_{j,i}$ ($j \in [1, M]$) are measured. All profiles are then preprocessed to temperature curves $\boldsymbol{C}_{j,i}^{k}$ according to the description in Section 3.1.
- *Step 2*: Under the definition of point pair in Section 2.2, the point $p_{j,i+1}$ is determined on the ($i$+1)-layer. Then its partial curve $\boldsymbol{C}_{j,i+1}^{k'}$ is predicted by the pretrained layer-to-layer prediction model $f_{s1}(\cdot)$, as discussed in Section 3.2.1.
- *Step 3*: The temperature profile of the point $p_{j,i+1}$ is constructed from the predicted curves $\boldsymbol{C}_{j,i+1}^{k'}$, as defined in Section 3.2.2. Then, the profile decomposition is performed to obtain the reduced basis $\widehat{\boldsymbol{U}}$ for the ($i$+1)-layer and the basis coefficients $\widehat{\boldsymbol{C}}_j$ for each point $p_{j,i+1}$. Based on the data from all $M$ points on the ($i$+1)-layer, an ELM model is trained online to learn the relationship between $\widehat{\boldsymbol{C}}_j$ and the relative delay $t_{rd}^{j}$ of point $p_{j,i+1}$.
- *Step 4*: Given a new point $p'$ on the ($i$+1)-th layer, its relative delay $t'_{rd}$ could be calculated from the travel speed and the distance to the boundary of the layer. Its basis coefficient $\widehat{\boldsymbol{C}}'$ is then predicted by the trained ELM model, and the temperature profile is constructed as the product $\widehat{\boldsymbol{U}}\widehat{\boldsymbol{C}}'$.

The key feature of the proposed online prediction method is that it relies on the temperature of one printed layer. If the thermal history of one layer could be measured during actual AM experiments, the online two-stage method could be applied to predict the thermal history of any points on the yet-to-print layer. However, this method is not applicable on the first layer of AM part, as no previous layer exists. Therefore, the thermal history of the first layer is assumed to be always measured in this paper.

## 4 Results and Discussions

Based on the collected experimental and simulation data, several steps are proposed to validate, and test the proposed online two-stage thermal history prediction method. Given the predicted profile of one point $p_{j,i}$ based on the predicted partial curves $\widehat{\boldsymbol{P}}_{j,i} = [\widehat{\boldsymbol{C}}_{j,i}^{1'}, \widehat{\boldsymbol{C}}_{j,i}^{2'}, \widehat{\boldsymbol{C}}_{j,i}^{3'}, \widehat{\boldsymbol{C}}_{j,i}^{4'}, \widehat{\boldsymbol{C}}_{j,i}^{5'}]$, and its actual profile from FEA simulations or WAAM experiments $\boldsymbol{P}_{j,i} = [\boldsymbol{C}_{j,i}^{1'}, \boldsymbol{C}_{j,i}^{2'}, \boldsymbol{C}_{j,i}^{3'}, \boldsymbol{C}_{j,i}^{4'}, \boldsymbol{C}_{j,i}^{5'}]$, the overall performance of the predicted profile is defined below, which is denoted as the relative error of profile (REOP),

$$REOP = \frac{1}{5N}\sum_{k=1}^{5}\sum_{n=1}^{N}\frac{\left|\widehat{T}_{j,i}^{k',n} - T_{j,i}^{k',n}\right|}{T_{j,i}^{k',n}}, i \in [1,35] \tag{13}$$

where $T_{j,i}^{k',n}$ is the $n$-th temperature data sampled on the actual curve $\boldsymbol{C}_{j,i}^{k'}$, and $\widehat{T}_{j,i}^{k',n}$ is the $n$-th predicted temperature data on the predicted curve $\widehat{\boldsymbol{C}}_{j,i}^{k'}$. In theory, a smaller REOP value indicates that the predicted profile matches the actual profile with smaller deviations. In this session, $N$ is set as 100 for all tests. Therefore, the input dimension and output dimension for the layer-to-layer prediction model $f(\cdot)$ is 104 and 100 respectively, which results in 1.8635 million parameters to be learned in the model. And only the performances of points on the former 35 layers are discussed, as they can provide the required five temperature curves. To reduce the effects of randomness, all tests are performed 10 times on a desktop computer with Intel(R) Core(TM) i7-3770 CPU @ 3.40GHz processor, 24.0 GB RAM, and 64-bit Windows 10 system. All prediction results and animations among 10 runs could be found here[†].

### 4.1 Performance of intra-layer prediction model

Before testing the entire online two-stage prediction method, the intra-layer prediction model is discussed first based on the profiles from simulation 1, as the overall performance of the layer-to-layer prediction model has been proved in our previous work [41]. For this purpose, the FEA profiles from points $p_{1,i}$, $p_{3,i}$, $p_{5,i}$, and $p_{7,i}$ on the $i$-th layer ($i \in [1,35]$) are selected to train the intra-layer prediction model. Then the profiles of points $p_{2,i}$, $p_{4,i}$, and $p_{6,i}$ are predicted and compared with the corresponding FEA profiles.

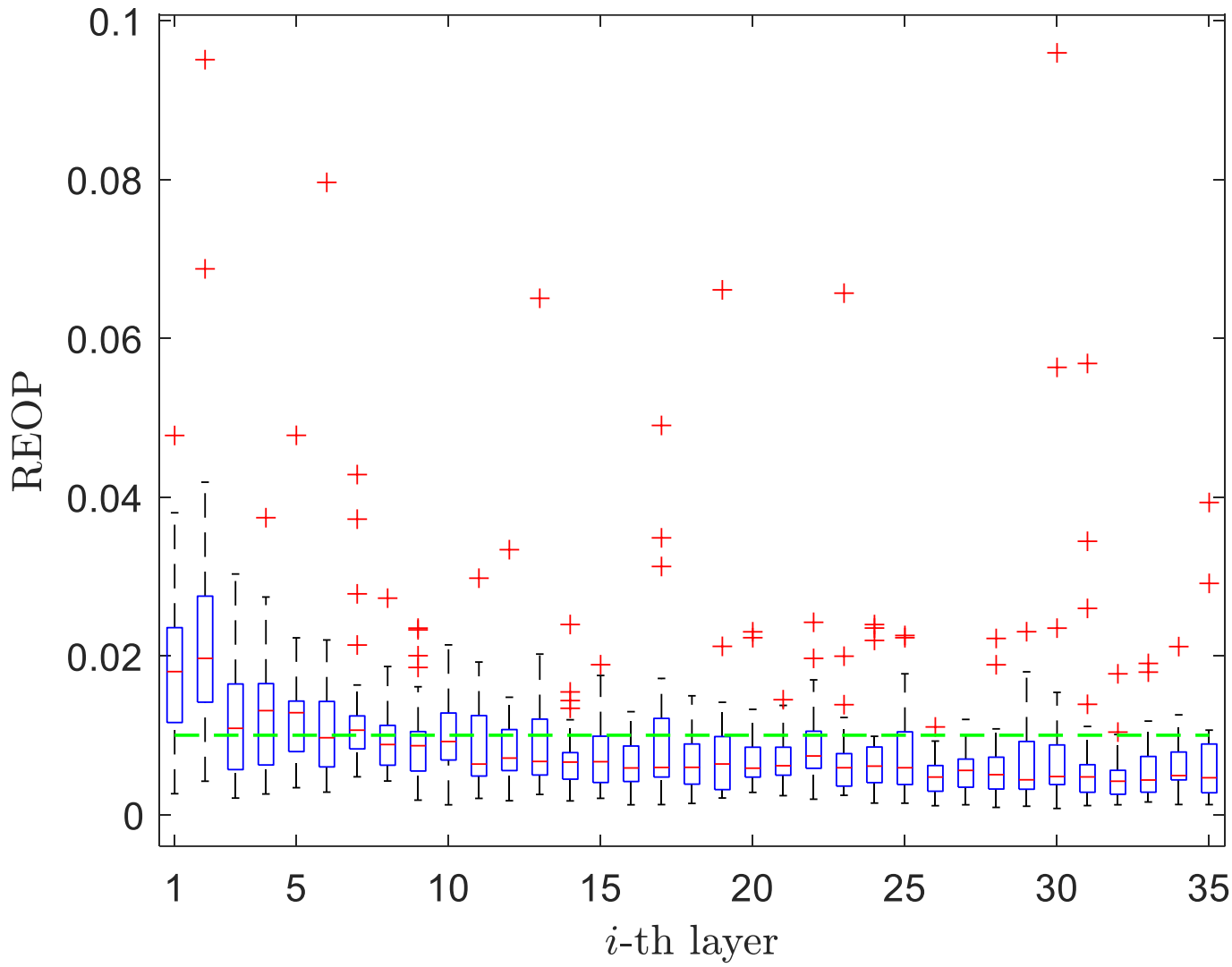

Figure 10 Boxplots of REOP values from points on different layers

The REOP values of predicted profiles on each layer among 10 runs are summarized in Figure 10, where two phenomena could be observed. First, most of the REOP values are smaller than 0.02, although some outliers are observed on each layer. As shown in Figure 11, the smaller the REOP value, the smaller differences are between the profiles. When the REOP value is around 0.1, the difference between the FEA profile and the prediction is observable visually. When the value is smaller than 0.03, the difference in profiles is visually undetectable, as shown in Figure 11 (d). Second, the overall prediction performance improves with the layer number. Below the fifth layer, the median REOP value of one layer is always larger than 0.01 (the green dashed line in Figure 10); above the fifth layer, the median REOP value is smaller than 0.01. Meanwhile, the size of the box also reduces with the increase of the layer number. According to Section 3.2.2, the performance of the proposed intra-layer prediction model (i.e., POD based ROM with ELM) is affected by the linearity of the profiles on the same layer. If the linearity is higher, the prediction performance would be better. In other words, the second phenomenon indicates that the linearity among temperature profiles also increases with the increasing layer number. One possible reason is that the thermal behavior is more stable on higher layers considering the heat accumulation on the lower layers. The above discussions demonstrate that the intra-layer prediction model enables the construction of the temperature profiles with acceptable accuracies.

---

[†] https://drive.google.com/file/d/1PN1ex1uaBMuLg8zWrE4LJMFt4KJ5XUZ4/view?usp=sharing

As the proposed prediction method is designed for online applications, the required CPU time for execution is another important factor. Given a trained layer-to-layer prediction model $f_{s1}(\cdot)$, the CPU time required for prediction increases with the number of testing curves, as shown in Figure 12 (a). In this paper, seven points are located on one layer and each point has five temperature curves. And only less than $0.01\ s$ is required to predict total 35 temperature curves from the curves on the lower layers. The execution time to construct the profiles of three points by the intra-layer prediction model $f_{s2}(\cdot)$ on each layer is summarized in Figure 12 (b), where the time on each layer is smaller than $0.02\ s$ in most cases. Therefore, the total execution time of the proposed online two-stage prediction method would be smaller than $0.1\ s$. Considering the dwell time of each layer varies from seconds to minutes during the completed CMT-WAAM experiments, the proposed prediction method is promising to be applied to the *in-situ* control or monitoring tasks.

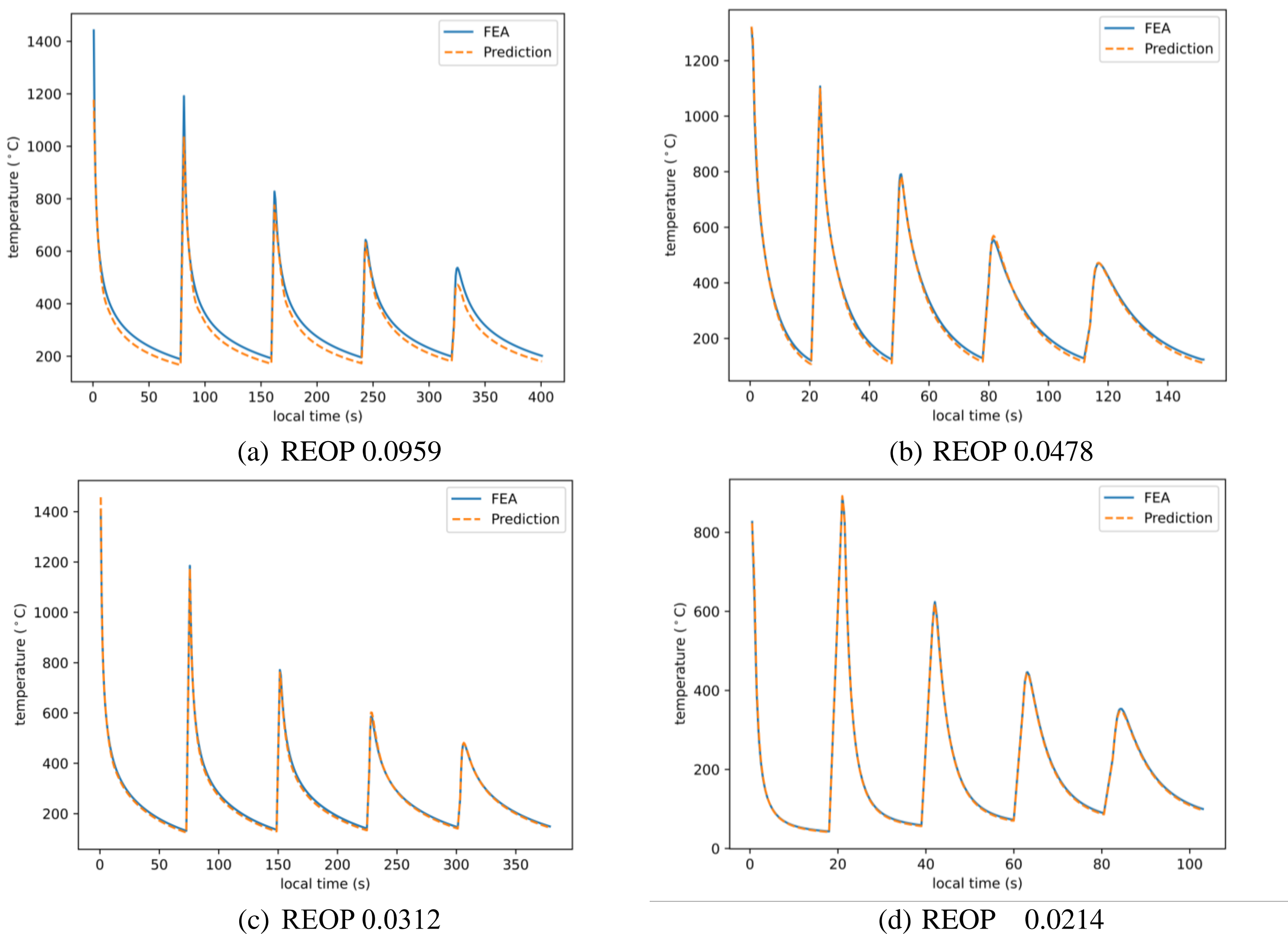

(a) REOP 0.0959

(b) REOP 0.0478

(c) REOP 0.0312

(d) REOP 0.0214

Figure 11 Comparison of FEA profiles and predicted profiles in testing $f_{s2}(\cdot)$ only

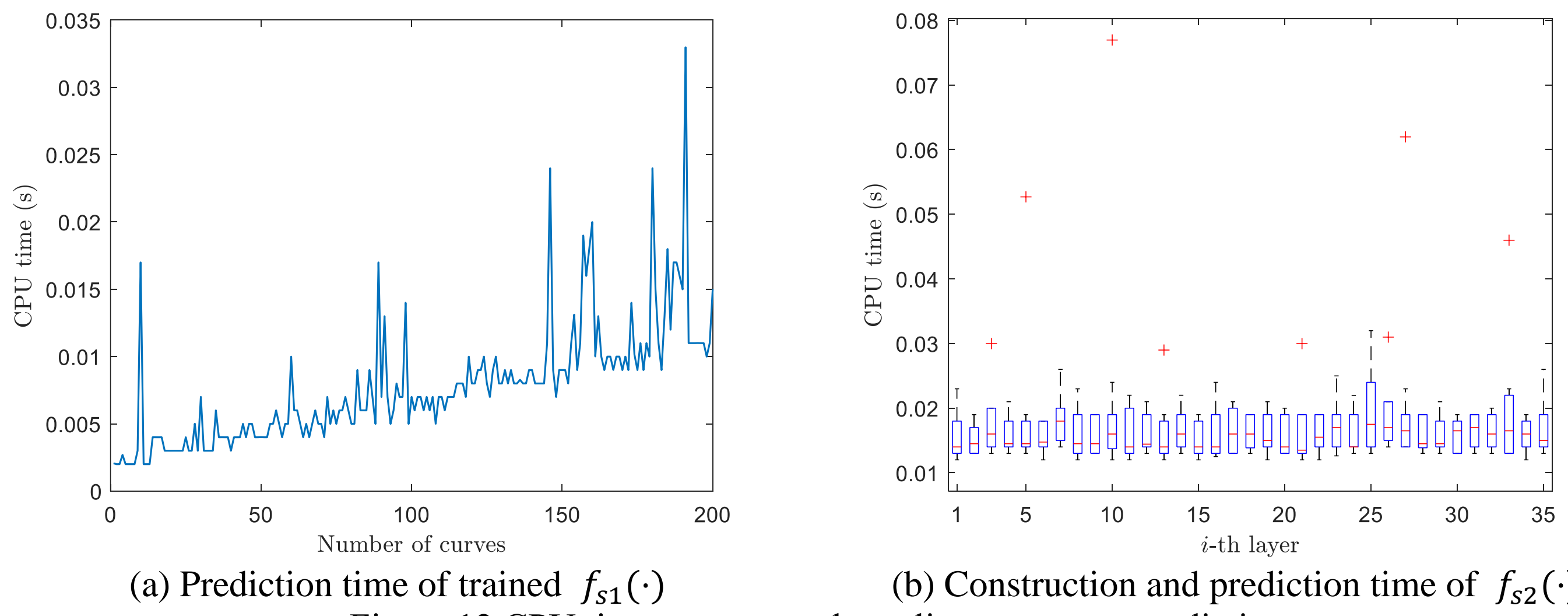

(a) Prediction time of trained $f_{s1}(\cdot)$ (b) Construction and prediction time of $f_{s2}(\cdot)$

Figure 12 CPU time to execute the online two-stage prediction

## 4.2 Validation with the same process setting

This section aims to check the generalization capability of the online two-stage prediction method within the same process setting. For this purpose, temperature profiles of the former 30 layers (i.e.,1,015 curve pairs) from Simulation 1 are used to train the layer-to-layer prediction model $f_{s1}(\cdot)$ first. Given the $i$-th layer ($i \in [31,35]$), temperature curves of points $p_{1,i}$, $p_{3,i}$, $p_{5,i}$, and $p_{7,i}$ are predicted by the trained model $f_{s1}(\cdot)$ from FEA curves of points on the ($i$-1)-th layer. Then the intra-layer prediction model $f_{s2}(\cdot)$ is applied to construct the temperature profiles of points $p_{2,i}$, $p_{4,i}$, and $p_{6,i}$, whose REOP values reflect the overall performances of the online two-stage prediction method on the $i$-th layer.

The boxplots of REOP values on each layer among 10 runs are shown in Figure 13. The median REOP values of all layers are smaller than 0.01, which indicates that the constructed temperature profile could match well the temperature profile from FEA in most cases, as shown in Figure 14 (c) and (d). When the outlier REOP value is smaller than 0.04, the predicted temperature profile is still acceptable, as only minor differences could be observed on profiles, as shown in Figure 14 (b). Although the outlier REOP value 0.1294 refers to an unacceptable profile with obvious deviations to the profile from the FEA simulation, the overall trend on the temperature profile is captured successfully, as shown in Figure 14 (a).

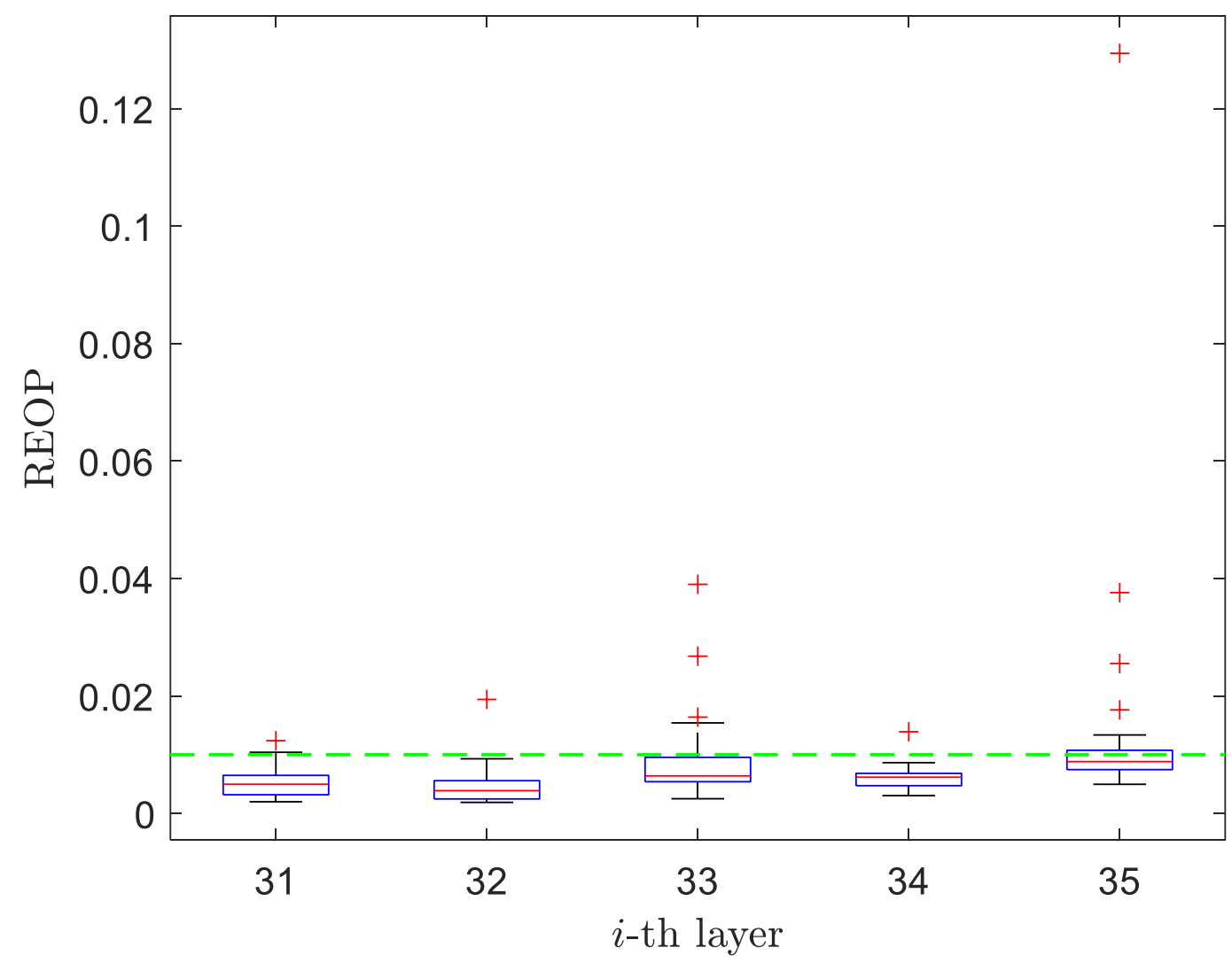

Figure 13 REOP of profiles on higher layers in validation with the same process setting

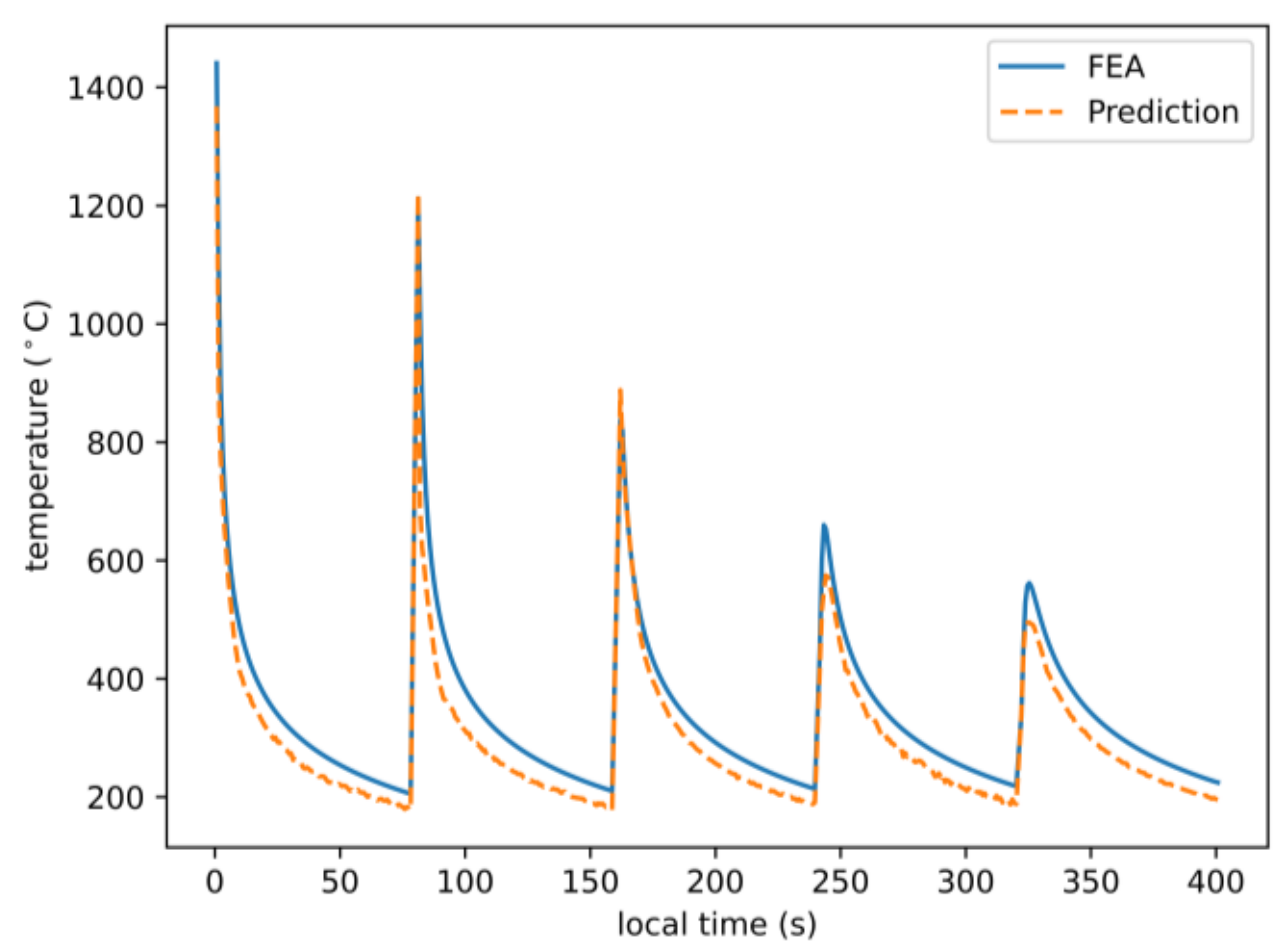

(a) Outlier REOP 0.1294

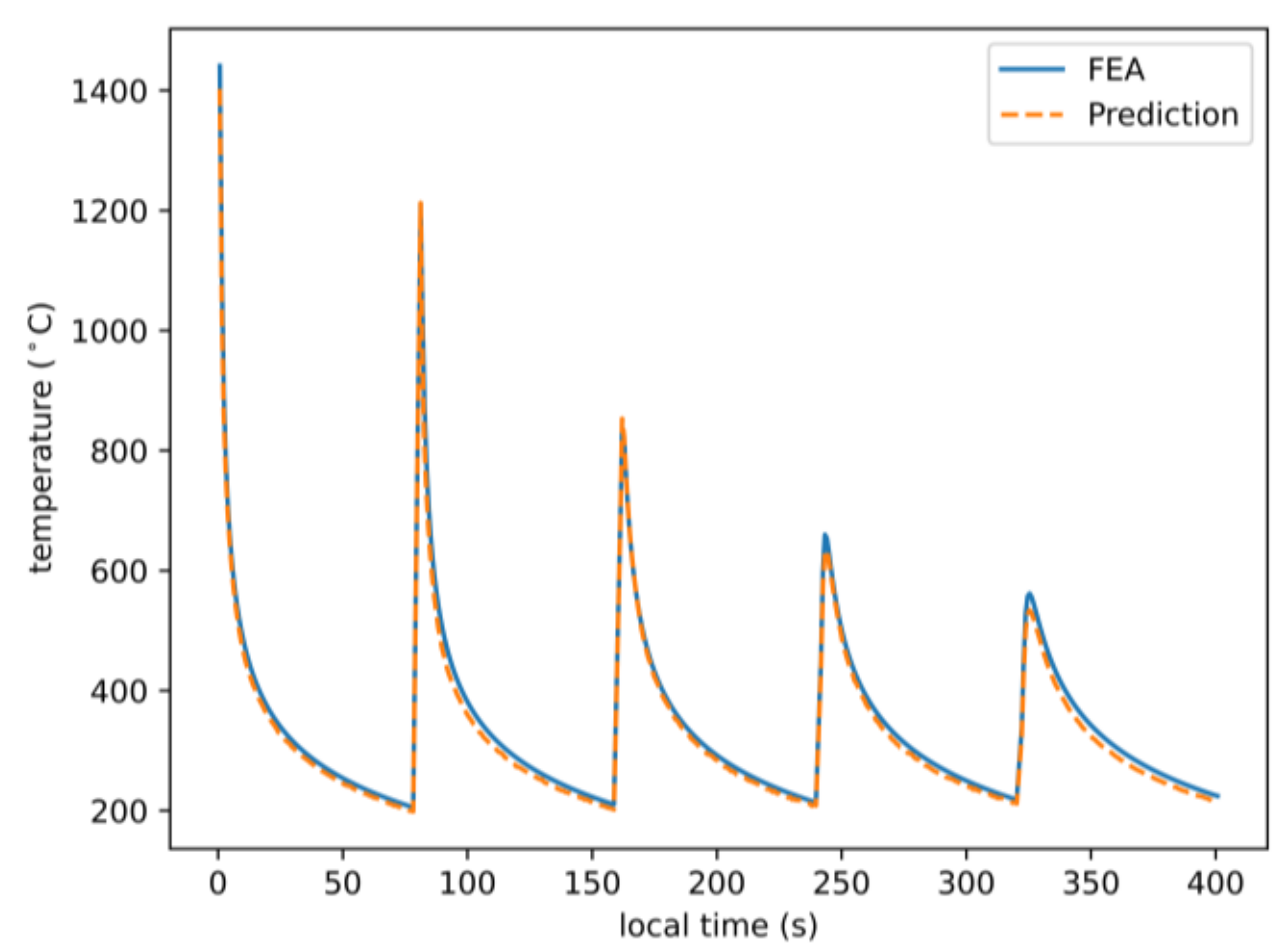

(b) Outlier REOP 0.0376

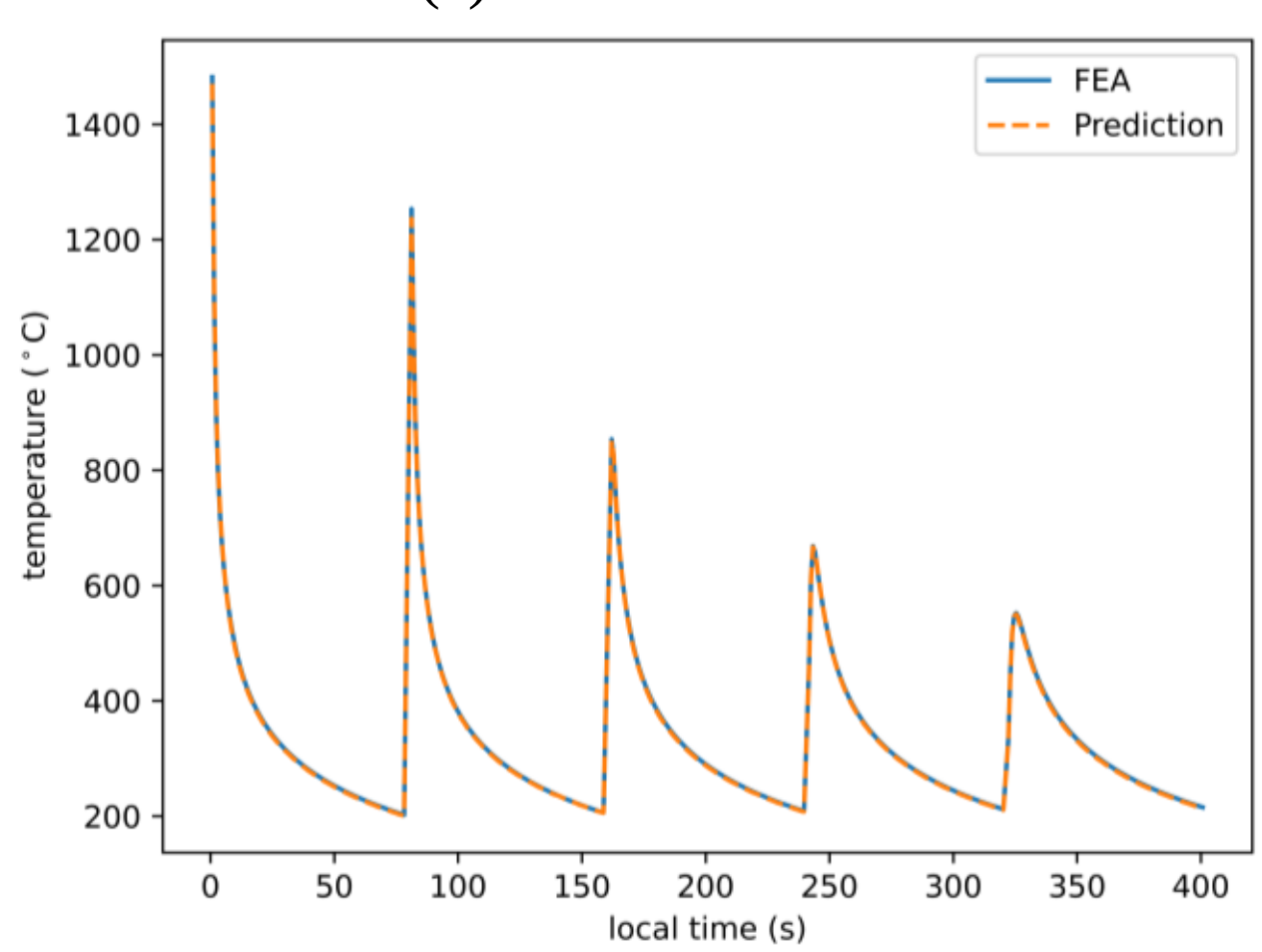

(c) Median REOP 0.0062

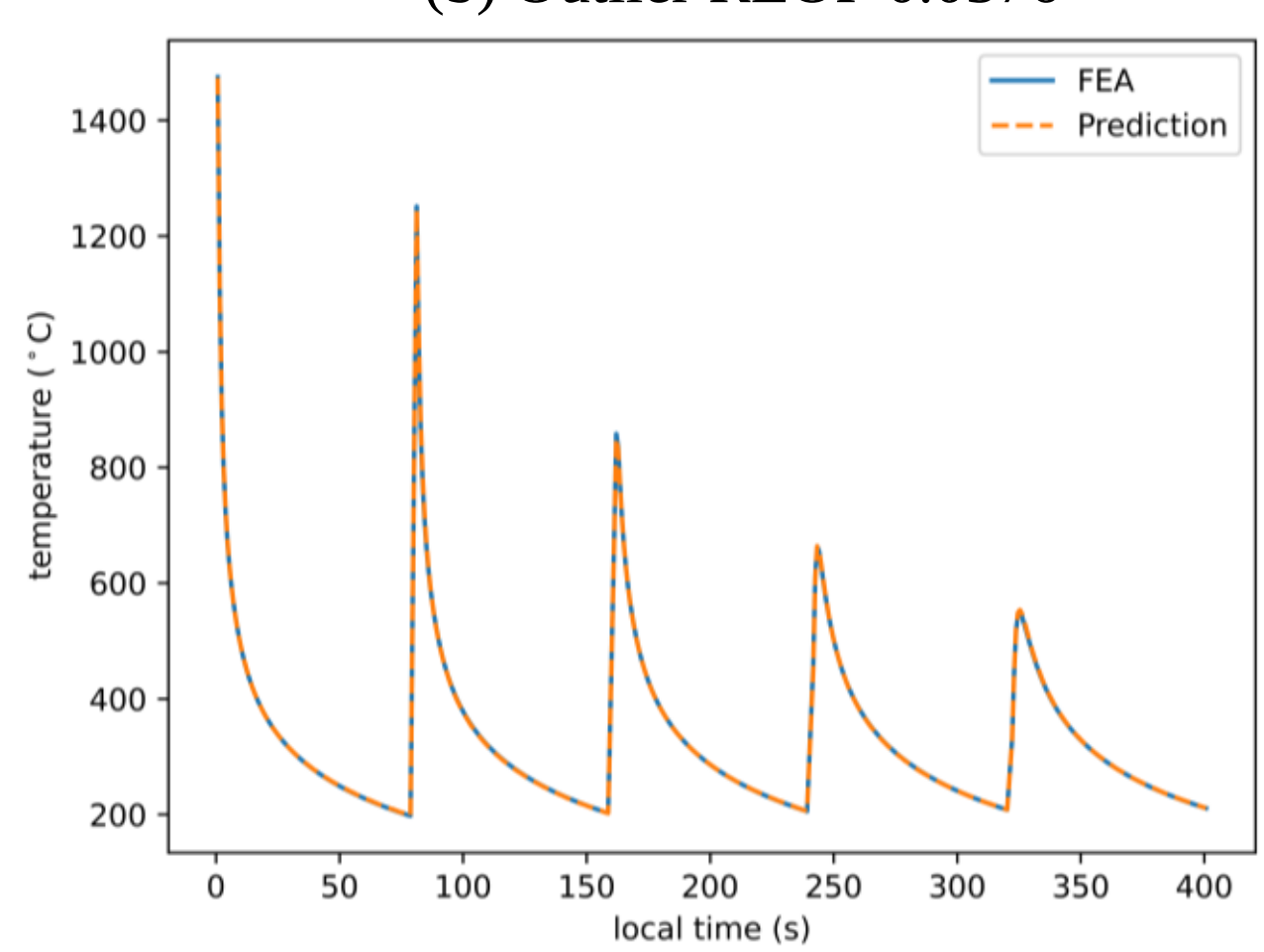

(d) Min REOP 0.0019

Figure 14 Comparison of FEA and predicted profiles in validation with the same process setting

Besides, to reflect the performance of the proposed method on constructing the thermal history of the entire layer, the intra-layer prediction model is used to construct the profiles of 160 points evenly located on the 31st layer, based on the FEA profiles and the predicted profiles of all seven points on the same layer. Then the temperature field is constructed according to the relative delays of all 160 points, and the temperatures of points which are not printed are assumed to be the room temperature. As only points located on the side of the layer are measured, the temperature inside the layer is not recorded in this work. For simplification, all points within the layer are assumed to behave the same along the width direction, i.e., holding the same temperature. Figure 15 depicts the temperature field of the 31st layer at different local times, where the time starting to print the layer is defined as the local time 0. From the temperature field at local time $6.0\ s$, some differences are observed near the boundary of the layer. The reason is that the boundary points of the layer are not covered to record temperature profiles, as shown in Figure 4. Then the prediction error could be amplified when extrapolating the temperature field near the boundary from the predicted profiles of points $p_{j,i}$ ($j \in [1,7]$). For the temperature fields at local times 19.5 $s$ and 93.0 $s$, the temperature fields of the layer constructed from both FEA and predicted profiles seem identical, especially in the region away from the boundary.

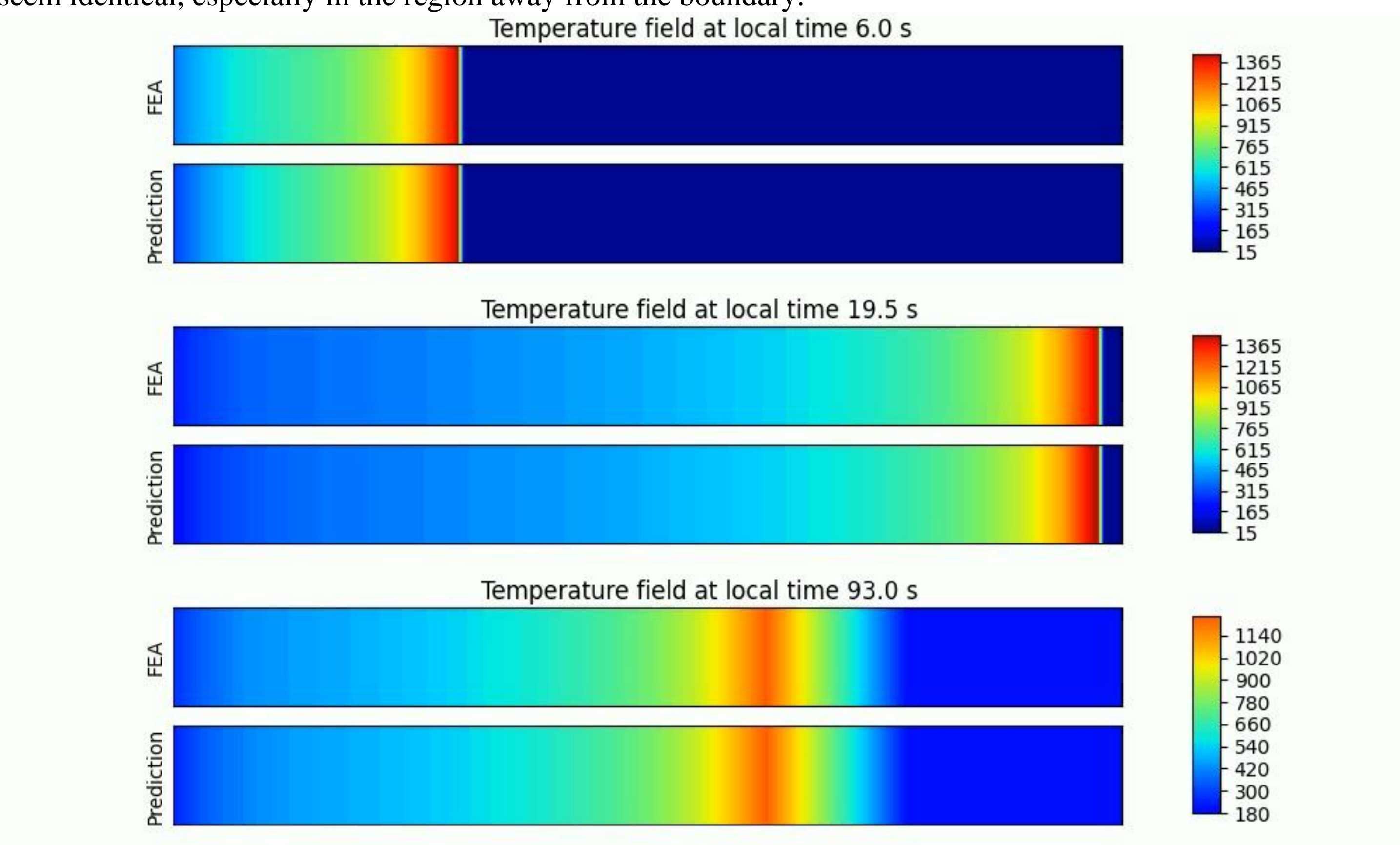

Figure 15 Temperature field of the 31st layer in validation with the same process setting

To sum up, the proposed online two-stage prediction method could predict the thermal history of the higher layer from the profiles of several points located on the lower layer, when the simulation settings (e.g., wire feed rate, travel speed, interpass temperature) of both training and testing data are the same. In other words, if the model could be trained based on profiles from the printed layers, the thermal history of the yet-to-print could be predicted with acceptable accuracies, i.e., most REOP values of predicted profiles are smaller than 0.01.

### 4.3 Validation with different process settings

In this section, the training and testing data are collected from different simulations, to study the generalization capability of the online two-stage prediction method among different process settings. More specifically, the layer-to-layer prediction model $f_{s1}(\cdot)$ is trained on the 10,080 curve pairs from Simulations 1-4 and Simulations 6-9. Then for Simulation 5, temperature profiles of points $p_{2,i}$, $p_{4,i}$, and $p_{6,i}$ on the $i$-th layer ($i \in [2,35]$) are constructed by the intra-layer prediction model from points $p_{1,i}$, $p_{3,i}$, $p_{5,i}$, and $p_{7,i}$, whose temperature profiles are predicted directly by the trained model $f_{s1}(\cdot)$ from the FEA profiles on the ($i$-1)-th layer. The REOP values at points $p_{2,i}$, $p_{4,i}$, and $p_{6,i}$ are selected to indicate the overall performance of the proposed method on the layer.

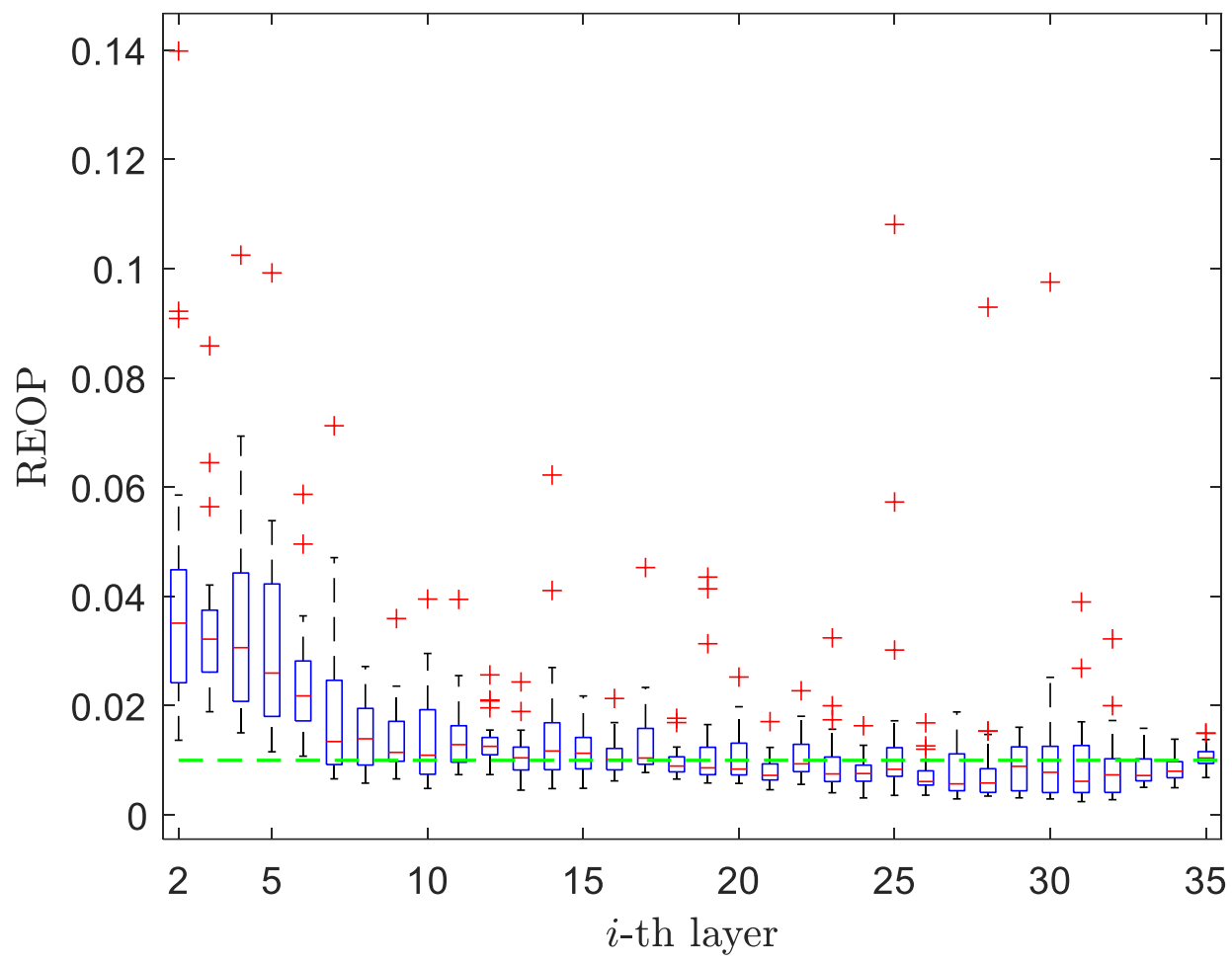

Figure 16 REOP of profiles on layers in validation with different process settings

Figure 16 shows the boxplots of obtained REOP values from the second to the thirty-fifth layer. Similar to the results in Figure 10, the overall performance of one layer increases with the height, which is attributed to the increased curve similarity between two successive layers (Section 3.2.1) and the increased linearity among temperature profiles on the same layer (Section 4.1). When testing the proposed method on the unseen simulation setting, the overall performance decreases compared with the performances in Sections 4.1 and 4.2. For example, the max outlier REOP value 0.1398 in the validation with different process setting is 46% larger than the max value in Figure 10. From the comparison in Figure 17 (a), the predicted profile has a large deviation on the peaks at the profiles. The maximum value at the first predicted curve is around 2200 °C while the value from the FEA model is below 1500 °C indicating an error over 700 °C. Besides, the box sizes from the 31st layer to the 34th layer in the validation with different process settings (Figure 16) are larger than the box sizes obtained in the validation with the same process setting (Figure 13). Although the median REOP values on the former five layers are larger than 0.02 in the validation with different process settings, all median REOP values are smaller than 0.01 above the 18th layer, and most REOP values are smaller than 0.02 above the eighth layer. The predicted profiles with the mean, the median, and the min REOP value are shown in Figure 17 (b) to (d) respectively, where the predicted profile still matches well with the profiles from the FEA simulation.

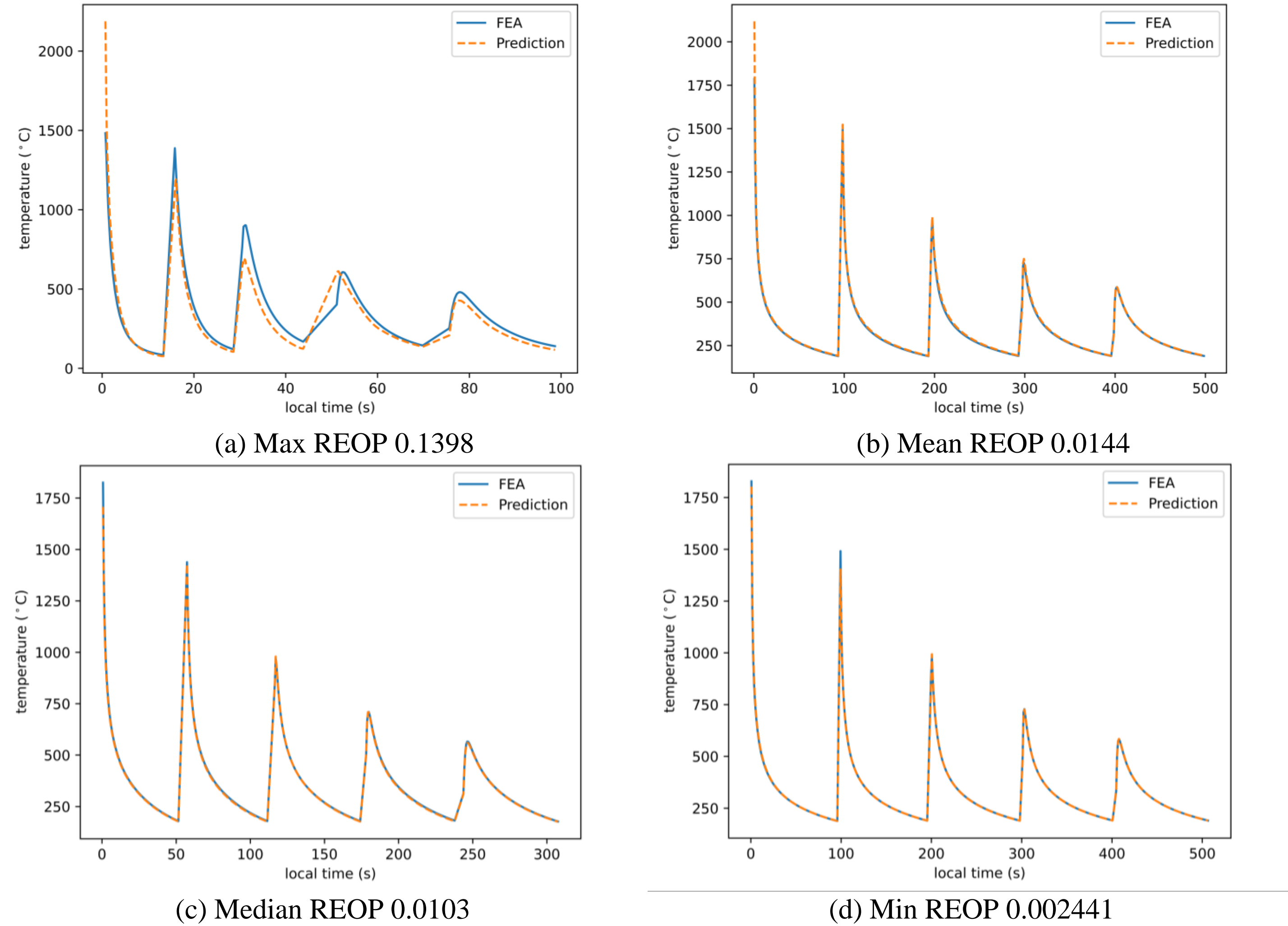

(a) Max REOP 0.1398

(b) Mean REOP 0.0144

(c) Median REOP 0.0103

(d) Min REOP 0.002441

Figure 17 Comparison of FEA and predicted profiles in validation with different process settings

Similar to Section 4.2, the FEA profiles and predicted profiles of points $p_{j,i}$ ($j \in [1,7]$) are used to construct the thermal history of the $10^{th}$ and the $20^{th}$ layer. In this section, only temperature fields of the $10^{th}$ and the $20^{th}$ layer at two different local times are shown in Figure 18 and Figure 19 respectively. Same as the observation in Section 4.2, the difference near the boundary also exists on both layers, as shown in the temperature field at local time 52.36 $s$ of the $10^{th}$ layer, and both temperature fields on the $20^{th}$ layer. Differently, the difference in the temperature field around the middle part of one layer is also obvious in the validation with different process settings. For instance, the predicted temperature field at local time 52.36 $s$ on the $10^{th}$ layer in Figure 18 has a wider part with high temperature than the FEA one. The same phenomenon is also applicable to the temperature field at local time 81.82 $s$ on the $20^{th}$ layer in Figure 19. Although some deviations are found in the temperature field, the trend of the entire layer is captured successfully on both layers, which indicates its capability to predict the thermal history of the higher layers in simulations with new process settings.

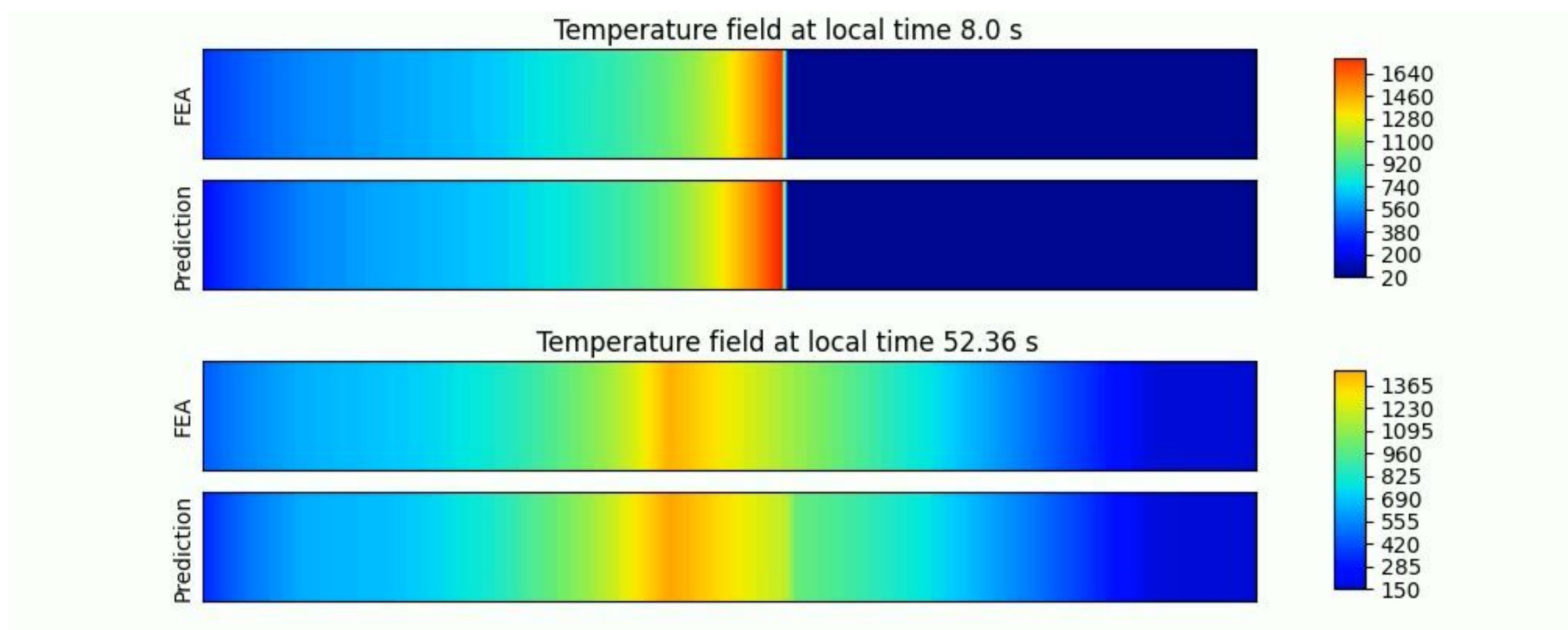

Figure 18 Thermal history of the 10th layer in validation with different process settings

Therefore, after training with sufficient data from simulations, the proposed online two-stage prediction method could be used to predict the thermal history of the higher layer when simulating the WAAM process with new parameters. This scenario is similar to online applications. If sufficient temperature profiles are collected from WAAM experiments to train the layer-to-layer prediction model, the online two-stage method could be used in other WAAM experiments with new process settings to predict the thermal history of the yet-to-print layer online, which is useful for *in-situ* monitoring and control.

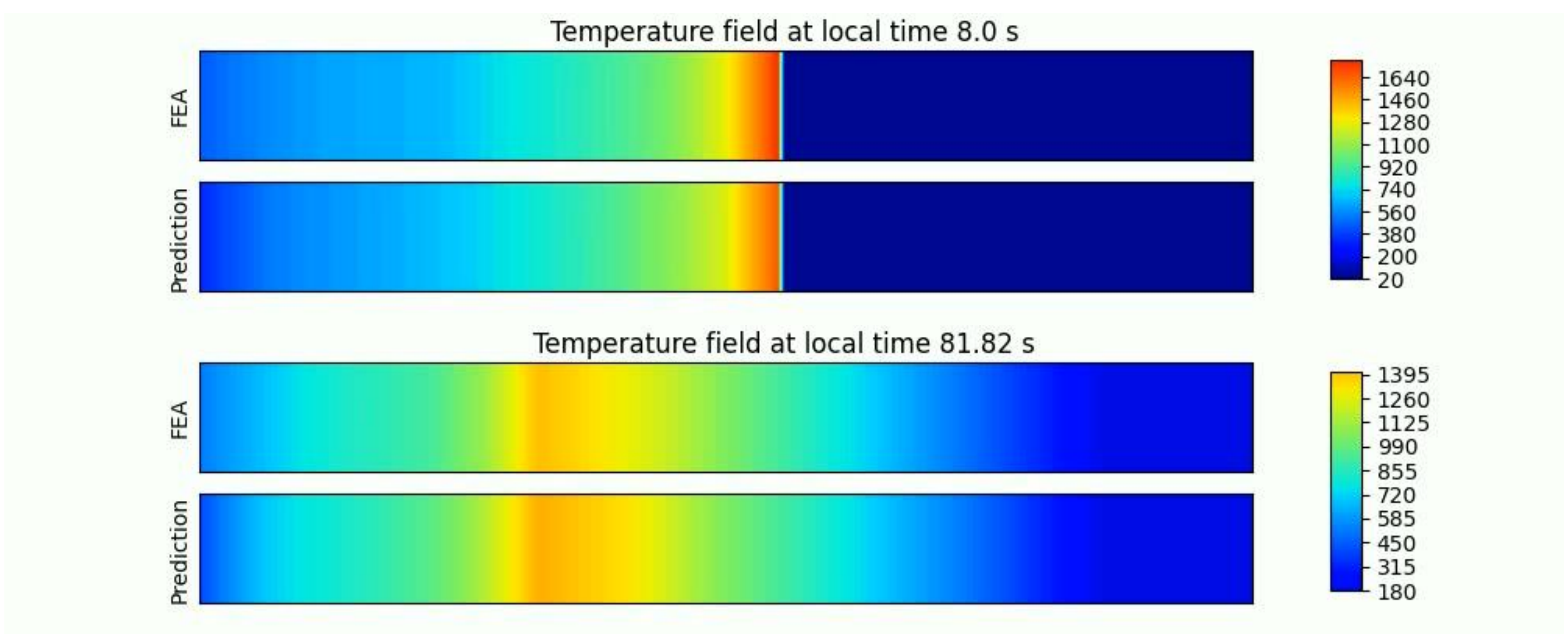

Figure 19 Thermal history of the 20th layer in validation with different process settings

## 4.4 Test of layer-to-layer prediction model

In this section, the intra-layer prediction model $f_{s2}(\cdot)$ is not discussed based on experimental profiles, as the thermal history of only one point on the layer is recorded during CMT-WAAM experiments, meaning that the profile decomposition by POD is not applicable anymore. As a result, only the layer-to-layer prediction model $f_{s1}(\cdot)$ is tested with experimental profiles. Besides, the model $f_{s1}(\cdot)$ is designed to learn the curve similarity between two successive layers, while the temperature profiles are measured at one point every five layers in the designed CMT-WAAM experiments as mentioned in Section 3.1.1. This indicates that it is not reasonable to apply the trained model $f_{s1}(\cdot)$ to the experimental curves directly. Therefore, a recursive prediction method is designed to predict the experimental temperature curves of the point on the ($i$+5)-th layer from the counterpart on the $i$-th layer, as shown in Figure 20. For instance, given one temperature curve $\boldsymbol{C}^{k}_{1,1}$ of the middle point on the first layer, the partial temperature curve of the middle point on the second layer is predicted as $\widehat{\boldsymbol{C}}^{k'}_{1,2} = f_{s1}(\boldsymbol{C}^{k}_{1,1}, t_{layer}, t^{1}_{dwell}, DR, h_1)$, based on which the corresponding partial temperature curve of the middle point on the third layer could be estimated as $\widehat{\boldsymbol{C}}^{k'}_{1,3} = f_{s1}(\widehat{\boldsymbol{C}}^{k'}_{1,2}, t_{layer}, t^{2}_{dwell}, DR, h_2)$. This prediction process continues until obtaining the partial curve of the middle point on the sixth layer, i.e., $\widehat{\boldsymbol{C}}^{k'}_{1,6} = f_{s1}(\widehat{\boldsymbol{C}}^{k'}_{1,5}, t_{layer}, t^{5}_{dwell}, DR, h_5)$. Then the REOP value of the predicted partial curve $\widehat{\boldsymbol{C}}^{k'}_{1,6}$ is calculated based on the experimental partial curve $\boldsymbol{C}^{k'}_{1,6}$ with the same time duration.

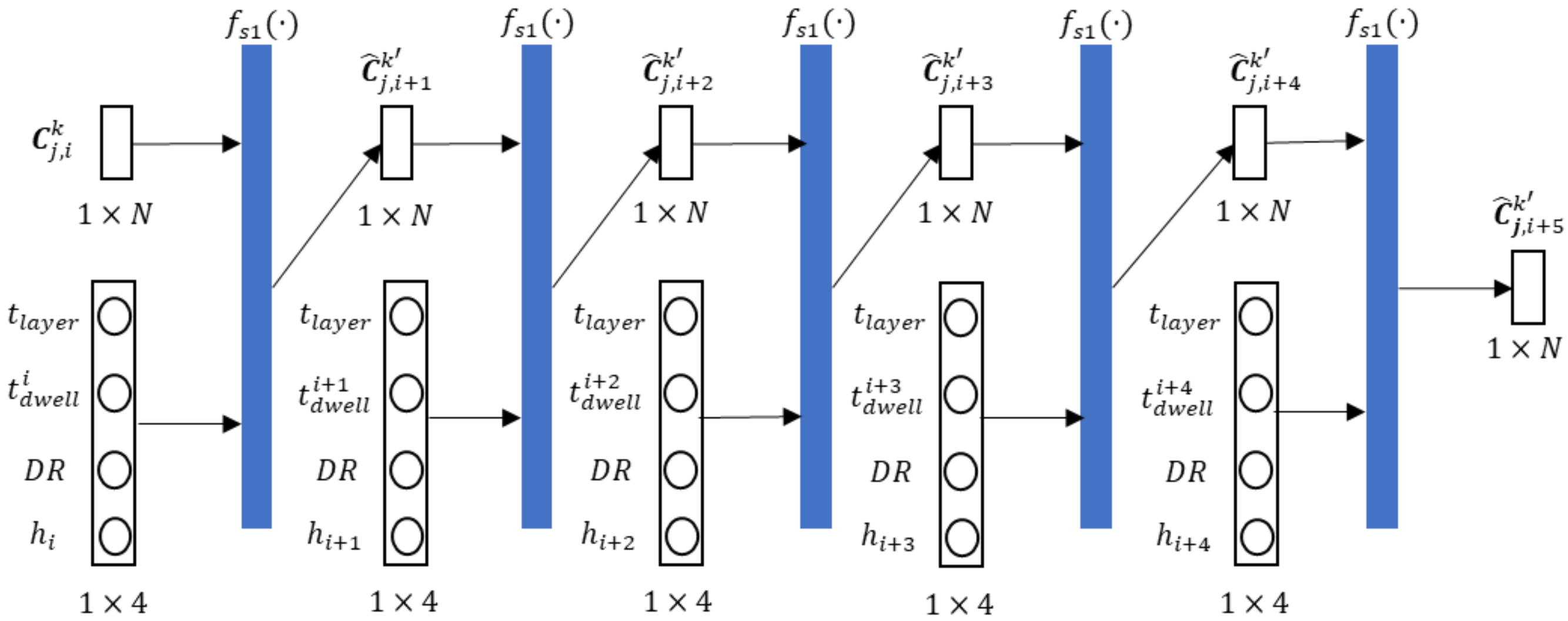

Figure 20 Recursive prediction method for experimental temperature curves

Theoretically, large prediction errors are anticipated if the designed recursive prediction method is tested on the experimental curves directly, considering the following two factors.

- *Simplified thermal behavior in simulations*. Generally, the thermal behavior in WAAM experiments is more sophisticated than the one in FEA simulations. Many assumptions have been applied in the FEA simulation for simplification, such as no geometry deformation, constant energy input, and fixed process parameters during printing. However, the geometry deformation and dynamic process parameters (e.g., travel speed, wire feed rate) during printing would make the experimental thermal behavior quite different from the one in simulations.
- *Error accumulation during recursive prediction*. In the proposed recursive prediction method, the prediction error on the partial curve $\widehat{\boldsymbol{C}}^{k'}_{j,i+1}$ would be passed to the next prediction $\widehat{\boldsymbol{C}}^{k'}_{j,i+2}$. Therefore, the error caused by the simplified thermal behavior would be amplified in the final prediction $\widehat{\boldsymbol{C}}^{k'}_{j,i+6}$.

To study the effect of the above two factors on the prediction performance, two test cases are defined for comparison.

- *Physics-uninformed prediction*: After training the layer-to-layer prediction model $f_{s1}(\cdot)$ with 11,340 temperature curves from all nine simulations, the trained model $f_{s1}(\cdot)$ is tested with temperature curves obtained from the eighth WAAM experiment. Therefore, no physical information about the actual experiment is contained in the trained model $f_{s1}(\cdot)$.

- *Physics-informed prediction*: After training $f_{s1}(\cdot)$ with all simulation data, the 115 temperature curve pairs from experiments 1-7 and experiments 9-15 are used to fine-tune all parameters in the trained model $f_{s1}(\cdot)$ based on the same training process in Section 3.2.1. As a result, the fine-tuned model learns some information about the physical thermal behavior in actual CMT-WAAM experiments.

The performance of both physics-uninformed and physics-informed predictions are summarized in Figure 21. In general, the performance of the physics-informed prediction outperforms the physics-uninformed prediction significantly. The REOP values of predicted profiles in the physics-uninformed prediction are larger than 0.1 in most cases, and the outlier REOP value could reach 0.75, as shown in Figure 21 (a). The predicted profile with the outlier REOP value is shown in Figure 21 (b), where the first and the second curves match the experimental curves well. But the third and the fourth curves have a large deviation, and the predicted temperature value is twice the experimental value. As a result, the profile obtained by the physics-uninformed prediction is unacceptable and cannot be applied in online applications.

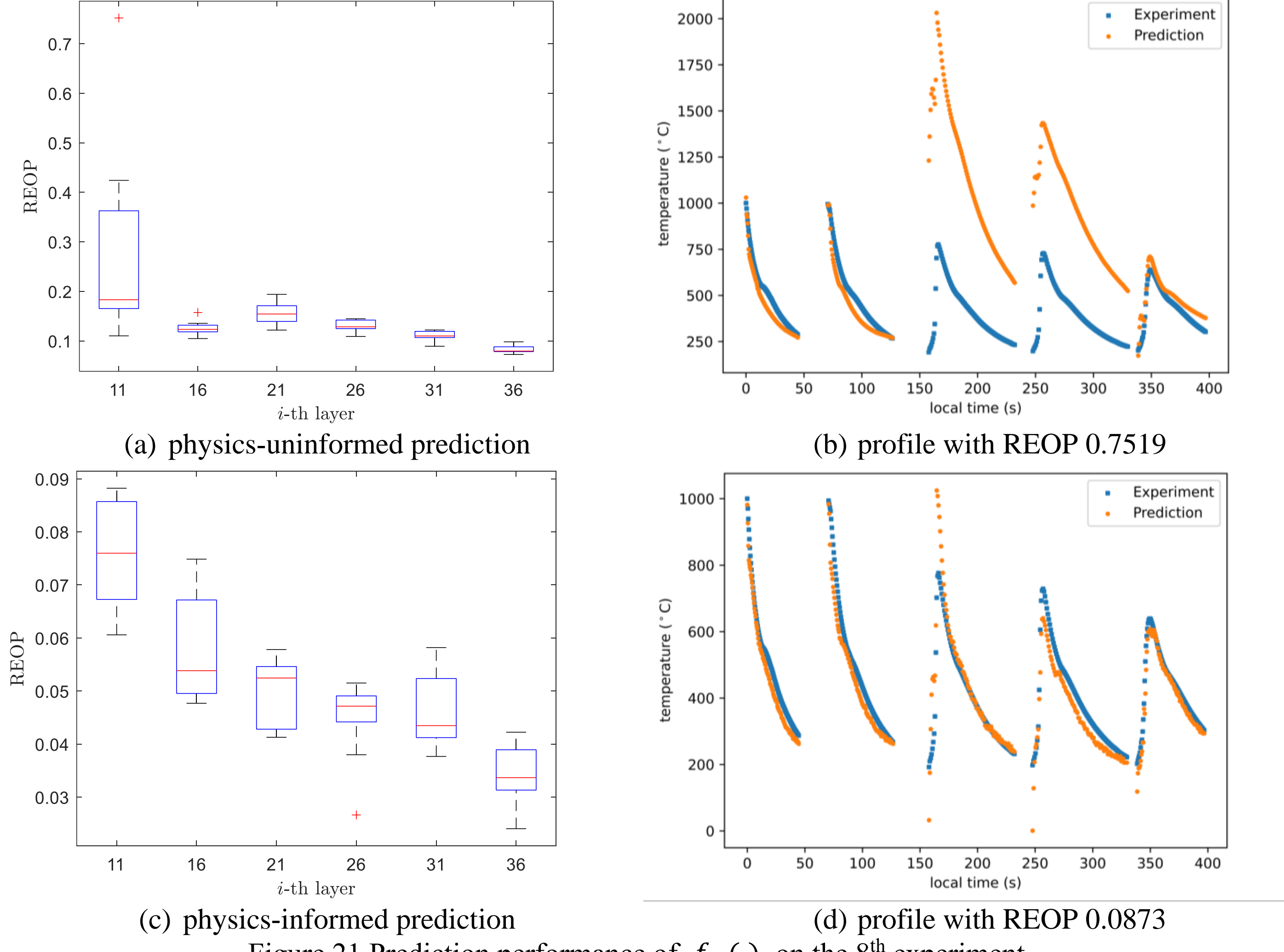

(a) physics-uninformed prediction (b) profile with REOP 0.7519

(c) physics-informed prediction (d) profile with REOP 0.0873

Figure 21 Prediction performance of $f_{s1}(\cdot)$ on the 8$^{th}$ experiment

In contrast, the REOP values of all predicted profiles from the physics-informed prediction are smaller than 0.09, as shown in Figure 21 (c). Same to the phenomenon in the FEA simulations, the prediction performance also increases with the layer number in experiments, which indicates that the experimental curve similarity increases with the height. Besides, the profile with the max REOP in the physics-uninformed prediction has a REOP value 0.0873 in the physics-informed prediction, as shown in Figure 21 (d). Compared with Figure 21 (b), all predicted temperature curves match the experimental ones better. This demonstrates that the proposed layer-to-layer prediction model could be applied in actual CMT-WAAM experiments after learning physical thermal behavior. Besides, compared with the training data size from simulations (i.e., 11,340), the size of experimental data (i.e., 115) for fine-tuning is quite small, which reflects the generalization of the proposed layer-to-layer prediction model in both simulation and experiments.

### 4.5 Summary remarks

Based on the above testing and discussion, three conclusions could be drawn about the proposed online two-stage prediction method: (a) The proposed method is efficient, as the execution CPU time to construct the thermal history of one layer from the temperature profile on the lower layer is less than 0.1 second on a commodity desktop computer (Intel Core i7-3770 CPU @ 3.40GHz processor, 24.0 GB RAM); (b) The proposed method has acceptable performances in simulation data, whether or not the training and testing data are from the simulation with the same process parameters; and (c) The proposed layer-to-layer prediction model can provide acceptable prediction performance on actual WAAM experiments after learning the physical thermal behavior from some experimental data. Therefore, we can claim that the proposed two-stage online prediction method is promising for online applications as far as the thermal history is concerned.

## 5 Conclusions

Different from current data-driven thermal history modeling methods whose offline performance is only tested with simulation data, this paper proposes a two-stage method for online thermal prediction, which is tested with both simulation and actual metal AM data. First, the similarities between temperature curves of any two successive layers (i.e., curve similarity) and temperature profiles of points on the same layer (i.e., profile similarity) are analyzed. Then, nine simulations in COMSOL Multiphysics software and fifteen wire arc additive manufacturing experiments are completed with corresponding profile processing methods. The online two-stage thermal history prediction method is developed based on the two observed similarities. Stage 1 aims to learn the curve similarity by a fully connected artificial neural network with residual connection, which is used to predict the temperature curves of points on the yet-to-print layer from the measured curves on the previously printed layer. Stage 2 designs an intra-layer prediction model leveraging the profile similarity. Given measured/predicted temperature profiles of several points on the same layer, a reduced order model (ROM) is built to predict the temperature of every point on the layer. The computationally efficient extreme learning machine is selected to train the ROM.

Based on the test results, three observations are made:

(a) The proposed online two-stage prediction method is computationally efficient since the time to construct the thermal history of one layer is within 0.1 seconds on a low-cost desktop computer.

(b) The proposed work has good generalization performance within the same simulation of the same AM process parameters and among simulations with different process parameters.

(c) The trained method shows a prediction performance with a relative error smaller than 0.1 on a new physical experiment, after fine-tuning with limited experimental data.

All test results demonstrate that the proposed online two-stage thermal history prediction method is promising for online thermal prediction and ultimately the control of metal AM processes.

For future work, the generalization performance on metal parts with more complex geometries could be studied if more pyrometers or thermal cameras are available to measure the thermal history online. The extension of the proposed method to other metal additive manufacturing processes, e.g., laser powder bed fusion, and direct energy deposition, could be explored.

## ACKNOWLEDGEMENTS

The authors gratefully acknowledge funding from the Natural Science and Engineering Research Council

(NSERC) of Canada [Grant numbers: RGPIN-2019-06601] and Business Finland under Project #: 4819/31/2021 with affiliation to the Eureka! SMART project (S0410) titled "TANDEM: Tools for Adaptive and Intelligent Control of Discrete Manufacturing Processes."